\documentclass{article}

\usepackage[preprint]{template/iclr2027_conference}
\usepackage{times}

\usepackage{amsmath,amsfonts,bm}

\def\eqref#1{equation~\ref{#1}}

\def\1{\bm{1}}

\DeclareMathAlphabet{\mathsfit}{\encodingdefault}{\sfdefault}{m}{sl}
\SetMathAlphabet{\mathsfit}{bold}{\encodingdefault}{\sfdefault}{bx}{n}

\usepackage{amsmath}
\usepackage{amssymb}
\usepackage{booktabs}
\usepackage{float}
\usepackage{placeins}
\usepackage{needspace}
\usepackage{graphicx}
\usepackage{wrapfig}
\usepackage{microtype}
\usepackage{multirow}
\usepackage{tabularx}
\usepackage{xcolor}
\usepackage{colortbl,enumitem}
\usepackage{hyperref}
\usepackage{url}
\usepackage{xurl}
\usepackage[nameinlink,noabbrev]{cleveref}
\hypersetup{
  hidelinks,
  pdftitle={PathFM: Fast Generative Forecasting of Multivariate Trajectories},
  pdfauthor={Jiachen Yao, Xiyuan Zhang, Annan Yu, Boran Han, Abdul Fatir Ansari, Shuai Zhang, Michael Bohlke-Schneider, Bernie Wang}
}

\definecolor{adapterteal}{HTML}{0F766E}
\definecolor{backboneblue}{HTML}{2563EB}
\definecolor{specialistorange}{HTML}{D97706}
\definecolor{softgray}{HTML}{F1F5F9}

\newcommand{\method}{PathFM}
\newcommand{\chronos}{Chronos-2}

\renewcommand{\eqref}[1]{\textup{(\ref{#1})}}

\title{\method{}: Fast Generative Forecasting of\\ Multivariate Trajectories}

\author{
Jiachen Yao\textsuperscript{1,}\thanks{Work done during an internship at Amazon.} \quad
Xiyuan Zhang\textsuperscript{2,}\thanks{Correspondence to: Xiyuan Zhang \textless\href{mailto:xiyuanz@amazon.com}{\texttt{xiyuanz@amazon.com}}\textgreater.} \quad
Annan Yu\textsuperscript{2} \quad
Boran Han\textsuperscript{2} \\
\bfseries Abdul Fatir Ansari\textsuperscript{2} \quad
Shuai Zhang\textsuperscript{2} \quad
Michael Bohlke-Schneider\textsuperscript{2} \quad
Bernie Wang\textsuperscript{2} \\[0.5em]
\normalfont\textsuperscript{1}California Institute of Technology \quad
\textsuperscript{2}Amazon
}

\begin{document}

\maketitle
\suppressfloats[t]

\begin{abstract}
Time series foundation models are becoming increasingly popular for general-purpose forecasting.
However, most leading models predict marginal quantiles, leaving the dependence structure of future trajectories unspecified.
This distinction matters for trajectory-dependent quantities such as cumulative demand, peak load, and event timing, yet point and marginal scores do not reveal it.
To model this dependence, we introduce \method{}, a trajectory-based forecasting model that uses flow matching to generate complete future paths.
We further distill the model to one or two steps and share history attention cache across draws.
Together, these enable \method{} to generate multiple paths for a single series in roughly the time needed for one query to a quantile model.
The distilled sampler also makes it practical to optimize a calibration loss directly.
\method{} achieves competitive marginal accuracy on the established forecasting benchmarks fev-bench and TIME.
To complement these evaluations, we introduce Forecast Path Score Bench (fps-bench), which augments fev-bench tasks with energy and variogram scores and downstream metrics.
\method{} achieves a higher aggregate path score on fps-bench than all evaluated external baselines.
Experiments on chaotic systems further show that our flow model sustains long-run variability, while strong baselines rapidly collapse.
\method{} makes generative time series forecasting both competitive and practical, enabling diverse downstream applications.
\end{abstract}

\begingroup
\setlength{\textfloatsep}{10pt plus 2pt minus 2pt}
\setlength{\floatsep}{8pt plus 2pt minus 2pt}
\setlength{\intextsep}{10pt plus 2pt minus 2pt}
\setlength{\abovecaptionskip}{6pt}
\setlength{\belowcaptionskip}{0pt}
\section{Introduction}
\label{sec:introduction}

Time series foundation models (TSFMs) are a hot topic~\citep{ansari2024chronos,das2024timesfm,ansari2025chronos2,timesfm3,cohen2025toto}.
The field is growing fast:
Every year, new models are released that refresh the leaderboard.
The models are commonly evaluated through point accuracy (e.g., MASE) and marginal probabilistic scores (e.g., CRPS, scaled quantile loss)~\citep{gift_eval,fevbench}.
Consequently, most of the recent TSFMs are designed and trained to produce quantile levels, which are optimized for these benchmarks.

Point accuracy and marginal probabilistic scores are widely adopted for good reason: they provide an interpretable, per-step summary of forecast quality that directly supports common operational decisions such as inventory planning, resource allocation, and risk management.
However, they do not capture all aspects of forecast quality.
Other properties, particularly \textit{trajectory-dependent} ones, are important but overlooked by these metrics.
Among many, we illustrate two important aspects:
\begin{itemize}[leftmargin=*,topsep=3pt,itemsep=3pt,parsep=0pt]
  \item \textbf{Learning distributions with multiple modes:} Quantiles are defined through the inverse of the cumulative distribution function (CDF) of the target distribution.
  These CDFs, however, are not always nice and smooth, for instance, when the target distribution has multiple modes or sharp peaks.
  With a few or a few dozens of quantiles, the resulting approximation of the CDF can be quite rough and suffer from a regression-to-the-mean issue~\citep{yu2026understanding}.
  See the Lorenz oscillator example in~\Cref{fig:ambiguity}.
  \item \textbf{Learning autocorrelation:} Many quantile regression models predict marginal quantiles at individual time steps across a forecast window.
  These marginal outputs do not specify temporal dependence among future values within that window.
  For instance, many forecasting problems also require learning trajectory-dependent quantities, such as cumulative demand, peak load, and threshold-crossing times \citep{perezdiaz2025form,cai2026procrusteanbedtimeseries}.
  See the solar-output example in~\Cref{fig:ambiguity}.
  A marginal quantile loss does not identify these joint properties.
\end{itemize}

\begin{figure}[htbp]
  \centering
  \includegraphics[width=\linewidth]{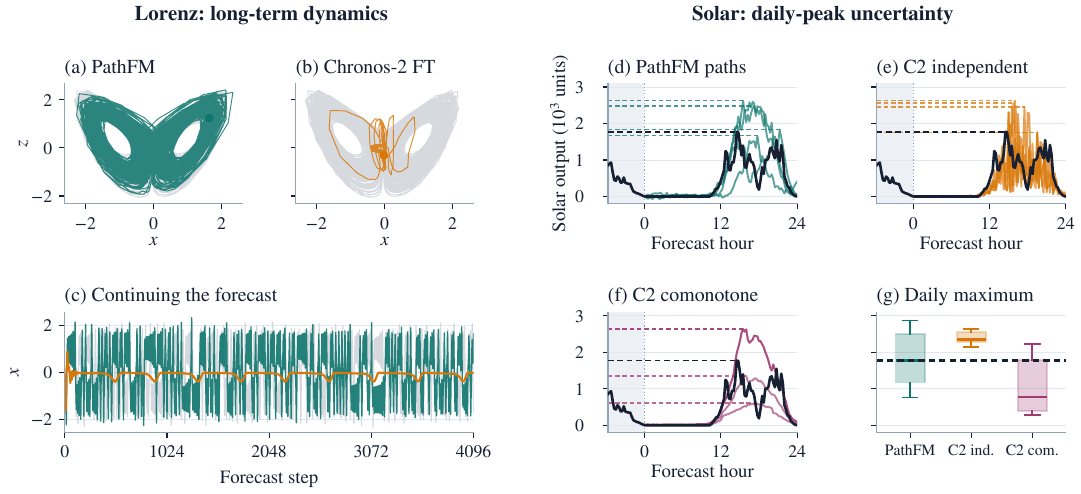}
  \caption{Two advantages of trajectory generation.
  Left: a sampled Lorenz rollout sustains variability and a two-lobe structure, while the fine-tuned Chronos-2 trace contracts.
  Both use the dynamical models in \Cref{sec:chaotic_systems}.
  Gray Lorenz curves show the observed continuation.
  Right: solar-output paths from \method{} and independent (ind.) and comonotone (com.) completions of the Chronos-2 (C2) quantiles.
  Horizontal dashed lines marks each solar path's daily maximum; black denotes the observed future.
  In (g), boxes show medians and interquartile ranges across 32 paths, whiskers show the 10--90\% range, and the dashed line marks the observed maximum.
  }
  \label{fig:ambiguity}
\end{figure}

Ad hoc methods have been proposed to mitigate these weaknesses, such as completing marginal quantiles under independence or comonotone assumptions~\citep{perezdiaz2025form}, or learning copula structures on top of marginal forecasts~\citep{drouin2022tactis,ashok2024tactis2}.
On the other hand, diffusion- or flow-matching-based generative time-series models are naturally well-suited for capturing joint trajectory distributions even without these ad hoc methods, as they can generate complete future paths while maintaining temporal and cross-variable dependencies.
Unfortunately, existing generative models~\citep{kollovieh2024tsflow,liu2025sundial} achieve neither high marginal accuracy nor efficient sampling, limiting their practical utility in real-world forecasting scenarios.
This leads us to ask:
\begin{center}
\setlength{\fboxsep}{0.01\linewidth}
\fcolorbox{black!30}{softgray}{\parbox{0.93\linewidth}{\centering\itshape
Can we construct a generative time-series model that achieves both high marginal accuracy and efficient sampling while also preserving trajectory dependence?}}
\end{center}

We introduce \method{} to combine joint and multi-mode trajectory modeling with accurate marginal forecasting and efficient sampling.
The model learns temporal and cross-variable dependence through conditional flow matching \citep{lipman2023flow}, generating complete future paths conditioned on history and known future covariates.
Parallel decoding distillation \citep{pdd2026} and history attention caching enable faster sampling with one or two model evaluations.
To narrow the marginal accuracy gap, we add a calibration loss that directly optimizes fair CRPS on generated forecasts.
The distilled sampler makes this optimization practical by allowing calibration gradients to pass through its short rollout and refine the forecast distribution.

With these components, \method{}-300M ranks \textit{second} on the widely used fev-bench for marginal scores (\Cref{sec:experiments}).\footnote{%
Cutoff date for the rankings throughout this paper: Sept.~26, 2026; see \Cref{app:concurrent-work} for concurrent work.}
On Forecast Path Score Bench (fps-bench), which we curate to evaluate trajectory quality, \method{} ranks \textit{first} for path scores (\Cref{sec:experiments}).
In a separate evaluation on chaotic systems, our sampling approach improves point forecast accuracy upon the state-of-the-art PANDA model, captures useful temporal dependence, and preserves long-run variability more faithfully than baselines (\Cref{sec:chaotic_systems}).
These results demonstrate that \method{} generates high-fidelity trajectory forecasts without sacrificing marginal accuracy, making it a strong candidate for forecasting applications.

Our contributions are as follows:
\begin{itemize}[leftmargin=*,topsep=3pt,itemsep=3pt,parsep=0pt]
  \item \textbf{Fast and calibrated trajectory generation.}
  We introduce \method{}, a joint flow foundation model for fast, calibrated trajectory generation with accurate small-ensemble marginals (\Cref{fig:training_pipeline}).
  Distillation enables one- or two-step generation, marginal-CRPS refinement improves empirical coverage, and stratified sampling reduces finite-ensemble marginal error.
  \item \textbf{A benchmark for trajectory distributions.}
  We introduce fps-bench, to our knowledge the first dedicated benchmark for evaluating path distributions from time series foundation models on real-world data.
  It reveals how strong marginal accuracy in existing models can mask errors in temporal dependence and uncertainty over trajectory-dependent quantities.
  \item \textbf{State-of-the-art performance.}
  \method{} outperforms leading foundation models on downstream probabilistic forecasting tasks and stays among the top on marginal benchmarks.
  It also achieves the highest forecasting accuracy across strong baselines on chaotic systems.
\end{itemize}

\section{Problem setup and preliminaries}
\label{sec:forecast_objects}

\paragraph{Forecast object and downstream query.}
Let $H\in\mathbb R^{T\times C}$ be the observed history and $Y\in\mathbb R^{L\times C}$ the future.
The available information $\mathcal I$ includes $H$, observation masks, and known future covariates.
A predictive law $P_\theta(Y\mid\mathcal I)$ supports queries about individual coordinates or a trajectory functional $g(Y)$, such as cumulative sum or autocorrelation.
A trajectory generator represents this law through $X=G_\theta(\mathcal I,Z)$, where each noise draw $Z$ produces one complete future path.

\citet{perezdiaz2025form} distinguish point, quantile, parametric-distribution, and trajectory forecast products by the decisions they support.
Coordinatewise quantiles $Q_\tau(Y_{\ell,c}\mid\mathcal I)$ and marginal distribution parameters leave temporal and cross-variable dependence unspecified.
Consequently, they do not determine predictive autocorrelation or distributions of sums, maxima, and path-dependent events without an additional dependence assumption.
Joint trajectory samples support direct estimation of those quantities by applying $g$ to each generated path.
\Cref{tab:model_capabilities} compares forecast objects, computational costs, and task compatibility; \Cref{sec:related-work} discusses related work.

\begin{table}[!htb]
  \centering
  \setlength{\abovecaptionskip}{0pt}
  \setlength{\belowcaptionskip}{5pt}
  \footnotesize
  \setlength{\tabcolsep}{3pt}
  \caption{Forecast objects, inference cost, and task compatibility of representative models.}
  \label{tab:model_capabilities}
  \newcommand{\taskyes}{\textcolor{green!50!black}{\ensuremath{\checkmark}}}
  \newcommand{\taskno}{\textcolor{red!75!black}{\ensuremath{\times}}}
  \begin{tabularx}{\linewidth}{llc
    >{\hsize=0.95\hsize\linewidth=\hsize\centering\arraybackslash}X
    >{\hsize=0.9\hsize\linewidth=\hsize\centering\arraybackslash}X
    >{\hsize=0.65\hsize\linewidth=\hsize\centering\arraybackslash}X
    >{\hsize=1.5\hsize\linewidth=\hsize\centering\arraybackslash}X}
    \toprule
    \multirow{2}{*}{Model} & \multirow{2}{*}{Forecast object} & \multirow{2}{*}{NFE} & \multicolumn{4}{c}{Task compatibility} \\
    \cmidrule(lr){4-7}
    & & & Pretrained & Marginal & Path & Cross-variable \\
    \midrule
    \rowcolor{gray!15}
    \multicolumn{7}{l}{\fontsize{8}{9}\selectfont\itshape\textcolor{black}{Quantile models}} \\
    Chronos-2 \citep{ansari2025chronos2}
      & Marginal quantiles
      & 1
      & \taskyes & \taskyes & \taskno & \taskno \\
    TimesFM-3 \citep{timesfm3}
      & Marginal quantiles
      & 1
      & \taskyes & \taskyes & \taskno & \taskno \\
    Toto 2.0 \citep{khwaja2026toto2}
      & Marginal quantiles
      & 1
      & \taskyes & \taskyes & \taskno & \taskno \\
    \rowcolor{gray!15}
    \multicolumn{7}{l}{\fontsize{8}{9}\selectfont\itshape\textcolor{black}{Flow models}} \\
    TSFlow \citep{kollovieh2024tsflow}
      & Joint paths
      & 32
      & \taskno & \taskyes & \taskyes & \taskyes$^\dagger$ \\
    Sundial \citep{liu2025sundial}
      & Univariate paths
      & 50
      & \taskyes & \taskyes & \taskyes$^\ddagger$ & \taskno \\
    \textbf{\method{} (Ours)}
      & Joint paths
      & 1--2
      & \taskyes & \taskyes & \taskyes & \taskyes \\
    \bottomrule
  \end{tabularx}

  \smallskip
  \begin{minipage}{\linewidth}
    \footnotesize
    \emph{Path} and \emph{Cross-variable} refer to native joint forecast laws.
    NFE counts evaluations per output block (per sampled path when applicable).
    $^\dagger$TSFlow's multivariate extension adds attention across features.
    $^\ddagger$Temporal coupling here refers to the block generated by the conditional flow.
  \end{minipage}
\end{table}

\paragraph{Existing benchmarks and their scope.}
GIFT-Eval reports point-forecast metrics and approximates univariate CRPS with weighted quantile loss \citep{gift_eval}; fev-bench uses scaled quantile loss across heterogeneous forecasting tasks \citep{fevbench}.
TIME evaluates zero-shot forecasting on freshly curated datasets, reporting MASE and quantile-based CRPS \citep{qiao2026time}.
These summaries assess central predictions and marginal uncertainty, including on multivariate tasks, but cannot distinguish joint laws with identical coordinatewise marginals.
We introduce fps-bench to address this gap by evaluating temporal dependence and trajectory-dependent quantities (\Cref{sec:fps_bench}).
\Cref{app:mc_limit} analyzes finite-ensemble approximation, and \Cref{app:metrics,app:chaos} define the evaluation metrics for general-corpus and chaotic-system prediction, respectively.

\section{Fast and Calibrated Trajectory Forecasting}
\label{sec:method}

\method{} learns a flow over complete future trajectories, then distills it for fast sampling and refines its marginal scores.
\Cref{fig:training_pipeline} summarizes the training and sampling stages.

\begin{figure}[!t]
  \centering
  \includegraphics[width=\linewidth]{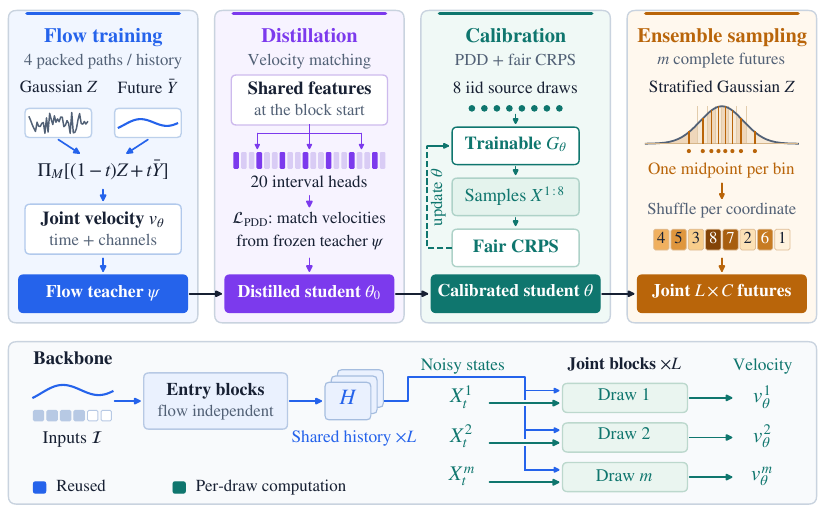}
  \caption{Training, post-training, and sampling with \method{}.
  Flow matching learns a joint velocity field; distillation matches frozen-teacher velocities at student-proposed states.
  Calibration maps eight iid draws through one fused step of trainable $G_\theta$ to samples scored by fair CRPS, together with PDD supervision.
  Sampling uses stratified initial noise to improve coverage of the noise distribution with a small ensemble.
  Below: Flow enters after the flow-independent prefix, and history attention values are computed once and reused by separate packed future states.}
  \label{fig:training_pipeline}
\end{figure}

\subsection{Pretraining by flow matching}
\label{sec:method_flow_matching}

Conditional flow matching~\citep{lipman2023flow} learns a velocity field that transports source noise toward the conditional law $P_\theta(Y\mid\mathcal I)$ over complete $L\times C$ futures.
Here $\mathcal I$ contains the observed history, masks, and known future covariates.
We transform targets to $\bar Y$ using context-derived location and scale followed by the inverse hyperbolic sine~\citep{ansari2025chronos2}.
We draw $Z$ independently of the future, with the same shape as $\bar Y$ and independent coordinates $Z_{\ell c}\sim\mathcal N(0,\sigma^2)$.
The mask $M$ selects valid unknown coordinates, $\Pi_M$ restores known values and zeros invalid coordinates, and $\|\cdot\|_M^2$ averages squared error over transported coordinates.

At flow time $t\in[0,1]$, we train $v_\theta$ on the straight path from noise to data:
\begin{align}
  X_t
    =\Pi_M\!\left((1-t)Z+t\bar Y\right), \quad
  \mathcal L_{\mathrm{FM}}
    =\mathbb E\left[
       \left\|v_\theta(X_t,t;\mathcal I)-(\bar Y-Z)\right\|_M^2
       \right].
  \label{eq:method_fm}
\end{align}
The expectation is over training pairs $(\mathcal I,\bar Y)$, source noise, and flow time.
Training uses one scalar flow time $t$ per row, shared across all channels and forecast steps (\Cref{app:representation}).

Inference integrates the ordinary differential equation, yielding $\widehat X_1$, one complete trajectory:
\begin{equation}
  \widehat X_0=\Pi_M(Z),\qquad
  \frac{\mathrm d\widehat X_t}{\mathrm dt}
  =M\odot v_\theta(\widehat X_t,t;\mathcal I),
  \quad t\in[0,1].
  \label{eq:method_flow_ode}
\end{equation}

\Needspace{7\baselineskip}
\subsection{Efficient backbone for joint generation}
\label{sec:method_backbone}

The velocity field uses a transformer based on T5~\citep{raffel2020t5}, with separate time and channel axes.
Its inputs are patches of 16 values with masks and time indices, a register token, and future patches populated by available covariates.
Each block combines temporal self-attention, attention across channels at aligned patches, and a pointwise feed-forward network.
We condition flow time via adaptive RMS normalization~\citep{chen2023pixart} and apply QK normalization and weight decay.

For reuse across paths, \emph{late entry} separates conditioning from flow-dependent computation.
The first $e$ blocks use unrestricted attention over history, the register, and future placeholders, without source noise or flow time.
These representations incorporate the requested horizon and known future covariates.
Restricting history attention before this stage would exclude that future information.
Future positions then receive a gated difference between flow-state and placeholder tokens.
In subsequent blocks, \emph{history-only self-attention} keeps history independent of the sampled path: history tokens attend only to history, while future tokens attend to both history and future.
Flow-time modulation acts only on register and future tokens.
\Cref{app:representation} specifies the configuration.

\paragraph{Packing and caching.}
Packed execution stores history once, with a separate register/future block per draw and masks preventing communication between paths.
Training averages four independent source/time draws per history, giving $4\times$ flow-matching supervision at $1.5\times$ the cost.
This sharing across paths is distinct from reusing history states across successive flow evaluations.
At inference, the first $e$ blocks run once per history and horizon; their outputs are reused across paths and flow evaluations.
The flow model can reuse deeper history keys and values across evaluations.
\Cref{sec:sampling_efficiency} compares accuracy and inference cost.

\subsection{Post-training: distillation with calibration}
\label{sec:method_posttraining}

Sampling a vanilla flow requires tens of sequential velocity evaluations, each depending on the state produced by the preceding step.
Distillation learns a shorter rollout, i.e., one or two evaluations.
This makes both ensemble generation and differentiation through generated forecasts practical.

\paragraph{Distillation.}
Parallel decoding distillation (PDD)~\citep{pdd2026} groups consecutive flow-time intervals into blocks, predicting all interval velocities in a block from one backbone evaluation.
We initialize the student from a flow teacher that remains frozen throughout post-training.
We divide $[0,1]$ into $J=20$ intervals with knots $\tau_j=j/J$ and widths $\Delta_j=\tau_{j+1}-\tau_j$.
Each interval has an affine output head $h_{\theta,j}=W_j\phi_\theta+b_j$.
All heads share features $\phi_\theta$ computed at the block's starting state and time, including nonlinear outputs and residual inputs.

For a block of $b$ intervals starting at $s$, the mean velocity advances $X$ to $X^+$:
\begin{align}
  \bar h_{\theta,s,b}
  =\sum_{j=s}^{s+b-1}
    \frac{\Delta_j}{\tau_{s+b}-\tau_s}\,h_{\theta,j}, \quad
  X^+=\Pi_M\!\left(X+(\tau_{s+b}-\tau_s)\bar h_{\theta,s,b}(X)\right).
  \label{eq:method_pdd_fusion}
\end{align}
Because the heads are affine in shared features, their interval-weighted parameter sums give a single readout per block.
One block of 20 intervals requires one evaluation; two blocks of 10 requires two.

For a selected interval $k$, summing preceding predicted displacements gives $\widetilde X_k$, where the teacher supplies head $k$'s target:
\begin{align}
  \widetilde X_k
  =\Pi_M\!\left(X_{\tau_s}
      +\sum_{j=s}^{k-1}\Delta_j\operatorname{sg}(h_{\theta,j})\right), \quad
  \mathcal L_{\mathrm{PDD}}
  =\mathbb E\left[
       \left\|h_{\theta,k}-\operatorname{sg}(v_\psi(\widetilde X_k,\tau_k;\mathcal I))\right\|_M^2
    \right].
  \label{eq:method_pdd_labels}
\end{align}
Here $\psi$ denotes the frozen teacher and $\operatorname{sg}$ stops gradients.
Gradients update the student through the selected head, with the proposed state and teacher target detached.
Targets are Euler velocities at student-proposed states, supplied without a complete teacher rollout.
Four packed draws each supervise one interval, with the flow-training coordinate normalization (\Cref{app:representation}).

\paragraph{Calibration.}
We refine the student's marginals by backpropagating the continuous ranked probability score (CRPS) through one fused sampling step.
Using few paths keeps this step affordable, but empirical CRPS has a bias that can favor underdispersion \citep{roberts2026ensembles}.
Fair CRPS~\citep{ferro2014fair} is unbiased for the marginal population score under iid sampling and finite first moments.
Let $G_\theta(Z;\mathcal I)$ denote the normalized endpoint of a fused block spanning $[0,1]$.
For $m$ iid source draws $Z^j\sim\mathcal N(0,\sigma^2 I)$, set $X^j=G_\theta(Z^j;\mathcal I)$.
At coordinate $(\ell,c)$, the fair estimator is
\begin{equation}
  \widehat C_{\ell c}
  =\frac{1}{m}\sum_{j=1}^{m}\left|X^j_{\ell c}-\bar Y_{\ell c}\right|
   -\frac{1}{2m(m-1)}
    \sum_{\substack{i,j=1\\i\ne j}}^{m}
      \left|X^i_{\ell c}-X^j_{\ell c}\right|.
  \label{eq:method_fair_crps}
\end{equation}
Retaining PDD provides teacher supervision while the generated endpoints are optimized against observations.
With $\widehat C_M$ the mean of $\widehat C_{\ell c}$ over transported coordinates, the objective becomes $\mathcal L_{\mathrm{cal}}=\mathcal L_{\mathrm{PDD}}+\mathbb E[\widehat C_M]$.
We evaluate trajectory quality using fps-bench (\Cref{sec:fps_bench}) and report direct calibration diagnostics in \Cref{app:calibration_controls}.

\subsection{Inference: ensemble sampling}
\label{sec:method_sampling}

At inference, \method{} generates $m$ complete future trajectories, from which we estimate quantiles and trajectory-dependent quantities.
We stratify the initial Gaussian noise to improve coverage at a fixed ensemble size, then propagate each draw through the distilled sampler.
\Cref{app:ensemble_sampling} constructs the ensembles.

\section{Evaluating Trajectory Forecasts}
\label{sec:fps_bench}

Trajectory forecasts should capture temporal dependence and accurately quantify uncertainty.
Marginal accuracy alone cannot establish this.
We introduce Forecast Path Score Bench (fps-bench) to evaluate these properties alongside marginal scores.
The benchmark combines established scoring rules into a common protocol for constructing and comparing finite trajectory ensembles.

\paragraph{Six complementary path scores.}
Different downstream queries depend on different aspects of a path, motivating the complementary diagnostics in \Cref{tab:score_objects}.
Following the aggregation-and-transformation framework of \citet{pic2025scores}, we score both complete paths and transformations that expose these aspects.
Energy score assesses the distribution of level paths, while differenced energy score emphasizes local changes by scoring the initial level and subsequent increments.
Variogram score compares pairwise temporal contrasts to probe how values vary together across time \citep{scheuerer2015variogram}.
We then apply scalar CRPS to the horizon total, maximum, and first-passage time to assess accumulated uncertainty, extremes, and event timing.
Specifically, first-passage time records the first crossing of a history-derived threshold.
fps-bench evaluates 94 fev-bench tasks \citep{fevbench}, excluding those with NaN values that prevent computing some path scores.

\paragraph{Quantile-model completions.}
\label{par:completions}
Applying these scores requires complete paths, which marginal quantile forecasts do not specify.
We retain native samples when available.
To include quantile models, we construct paths from their marginal quantiles under two explicit dependence assumptions.
Sorted quantiles define a piecewise-linear inverse CDF with constant extension beyond the outer quantiles.
An independent completion draws a separate quantile rank at each forecast time and channel, whereas a comonotone completion shares one rank across an item's times and channels.
Both laws preserve the same underlying interpolated marginals, allowing us to examine how the dependence assumption affects trajectory scores.
These two choices serve as baselines, as different dependence assumptions may yield different scores for the same marginal forecasts.
\Cref{app:metrics} specifies the score definitions, estimators, and aggregation.

\section{Forecasting accuracy and efficiency}
\label{sec:experiments}
\label{sec:results}

We evaluate forecasts through their marginal accuracy, the path quality scores, and the cost of generating an ensemble.
The experiments establish where trajectory forecasts help, then examine how distillation and score refinement make those forecasts practical.
\Cref{sec:chaotic_systems} complements these experiments with a study of chaotic systems, with a particular focus on long-run dynamics.

\Needspace{7\baselineskip}
\subsection{Experimental setup}
\label{sec:experimental_setup}

Unless stated otherwise, \method{} refers to the 126M model; we use \method{}-126M and \method{}-300M when comparing sizes.
We evaluate marginal accuracy on fev-bench \citep{fevbench} and TIME \citep{qiao2026time}.
Baselines include TimesFM-3 \citep{timesfm3}, \chronos{} \citep{ansari2025chronos2}, Toto-1.0/2.0 \citep{cohen2025toto,khwaja2026toto2}, and TiRex/TiRex-2 \citep{auer2025tirex,podest2026tirex2}.
Quantile models use explicit independent or comonotone completions, whereas Toto-1.0 and Sundial \citep{liu2025sundial} retain their native sampled paths.
GIFT-Eval results appear in \Cref{app:benchmark_transfer}, and the full component comparisons in \Cref{app:benchmark_details}.

\method{} and all fps-bench forecasts use 32 paths per forecast.
For \method{}, we use stratified Gaussian sources and one model evaluation per output block.
For each quantile model, we evaluate independent and comonotone completions and report the higher aggregate skill.

\subsection{Results}

\begin{figure}[t]
  \centering
  \includegraphics[width=\linewidth,trim=0 4bp 0 0bp,clip]{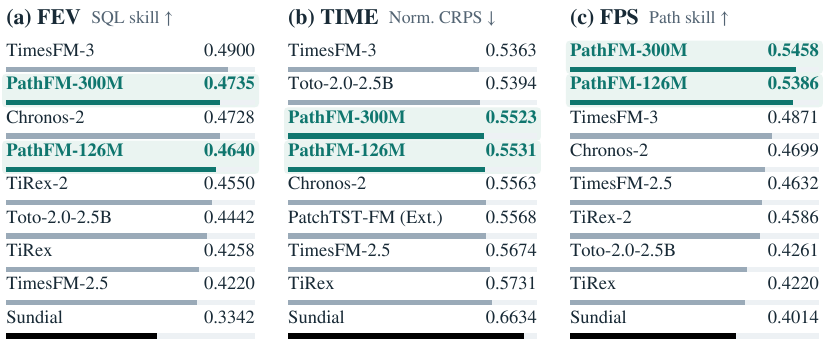}
  \caption{Marginal accuracy and trajectory quality for \method{}-126M and \method{}-300M.
  Native samplers \method{} and Sundial are teal and black.
  FPS selects each quantile model's better independent or comonotone completion.
  }
  \label{fig:benchmark_comparison}
\end{figure}

\method{}-300M and \method{}-126M achieve the two highest FPS scores among the compared models while remaining competitive on marginal accuracy (\Cref{fig:benchmark_comparison}).
The 300M model ranks second on fev-bench and third on TIME; both sizes have lower TIME CRPS than \chronos{}.
On fps-bench, both outperform TimesFM-3 and \chronos{} even when each quantile baseline uses its better independent or comonotone completion.
Both also improve on Sundial's native sampled forecasts across all three benchmarks.
TimesFM-3 leads both marginal benchmarks, but its lower FPS ranking illustrates why marginal accuracy alone is insufficient to assess forecast paths.

\paragraph{Path-dependent scores.}
\Cref{fig:fps_component_bars} compares the six FPS components for \method{} and the three strongest external baselines.
The clearest gains over TimesFM-3 (the strongest baseline) occur in scores for temporal changes, horizon totals, and maxima.
These gains highlight the value of complete paths for downstream tasks that depend on temporal structure beyond marginal forecasts.

\paragraph{Ablation studies.}
We study the cumulative effects of architecture and learning choices in \Cref{app:ablation}.
The teacher--student evaluation-budget comparison (\Cref{app:pdd_nfe}) and inference speedup breakdown (\Cref{fig:sampling_tradeoffs}) examine how distillation and shared history computation reduce sampling cost.
\Cref{app:scaling} provides additional results on GIFT-Eval.
We also evaluate cross-variable dependence (\Cref{app:cross_variable}) and test fitted temporal copulas for quantile baselines (\Cref{app:temporal_copulas}).

\Needspace{7\baselineskip}
\subsection{Accuracy, model size, and inference cost}
\label{sec:sampling_efficiency}

\method{} combines the highest FPS skill among the compared models with a compact 126M-parameter network (\Cref{fig:model_family_latency}).
With one evaluation per output block, it generates 32 paths in 51.6\,ms on an NVIDIA L4, placing it on the observed accuracy--latency frontier for path forecasts.
We compare the number of function evaluations in \Cref{app:pdd_nfe} and sampling controls in \Cref{app:latency}.

\begin{figure}[t]
  \centering
  \includegraphics[width=\linewidth]{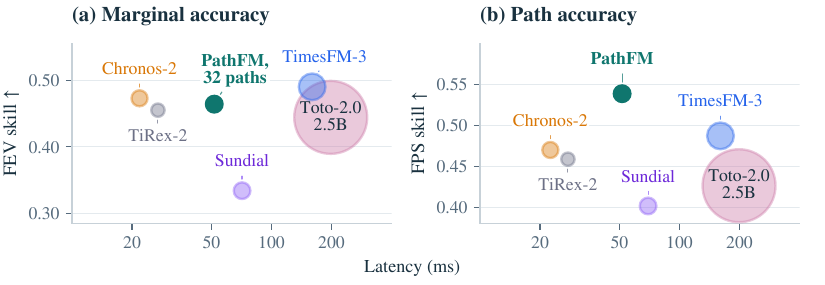}
  \caption{Accuracy, latency, and model size across six forecasting models.
  Area denotes parameter count.
  Latency is the median time to return nine quantiles (a) or forecast paths (b) per query on an NVIDIA L4, with history and forecast lengths fixed across models.
  Quantile models use their better aggregate completion from \Cref{par:completions}.
  }
  \label{fig:model_family_latency}
\end{figure}

\begin{figure}[t]
  \centering
  \includegraphics[width=\linewidth]{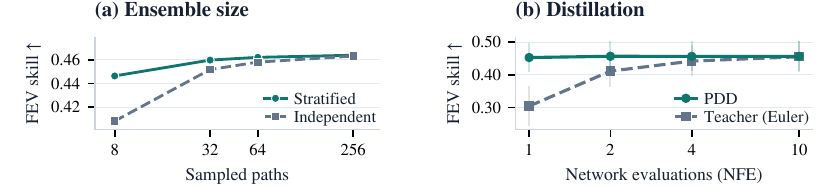}
  \caption{Ensemble size and distillation on fev-bench tasks.
  (a) Calibrated 126M model with different sampling strategies at NFE=1 (\Cref{app:ensemble_sampling}).
  (b) Its matched teacher and PDD student.
  Thin bars in (b) show pointwise 95\% task-bootstrap intervals (\Cref{app:pdd_nfe}).
  }
  \label{fig:sampling_distillation}
\end{figure}

\paragraph{Ensemble size and distillation.}
\Cref{fig:sampling_distillation} examines two ways to reduce the sampling budget in separate controlled studies.
On all 100 fev-bench tasks, stratification improves marginal accuracy most at small ensemble sizes: 32-path stratified sampling achieves almost the same FEV skill as 256 iid paths.
With one evaluation, the student approaches the ten-evaluation teacher; two evaluations slightly exceed its FEV skill.
This shows the power of distillation for time series flow matching models, and the effectiveness of the parallel decoding distillation.

\paragraph{Scaling to 300M parameters.}
We investigate scalability by comparing \method{}-126M with \method{}-300M.
\Cref{fig:training_scaling_curves} tracks flow pretraining; the larger model reaches lower training and corpus-evaluation losses at the same number of updates.
The larger model improves both marginal accuracy and trajectory quality (\Cref{fig:benchmark_comparison,app:scaling}).
These gains come at roughly twice the inference latency for generating 32 paths on an NVIDIA L4 (\Cref{fig:pathfm_scaling_latency}).
The scaling curves suggest that larger models may continue to improve accuracy.
This shows the potential for larger flow models.

\begin{figure}[t]
  \centering
  \includegraphics[width=\linewidth,trim=0 9bp 0 0bp,clip]{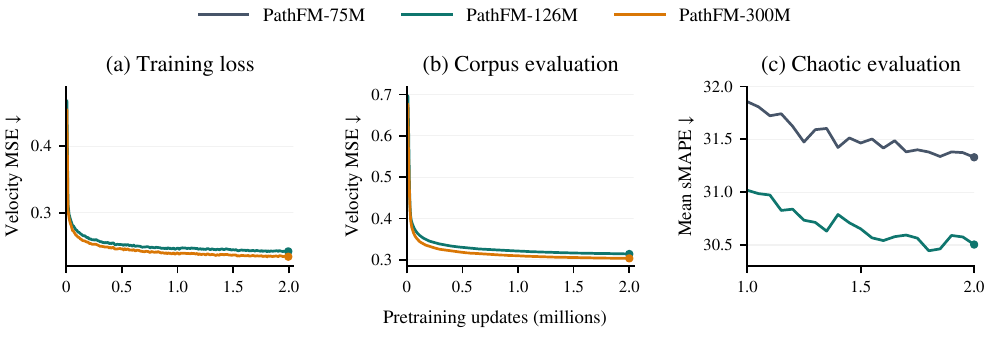}
  \caption{Pretraining progress on general and chaotic time series.
  (a) Training velocity MSE; (b) EMA velocity MSE on fixed corpus windows for \method{}-126M and \method{}-300M;
  (c) EMA mean validation sMAPE for the 75M and 126M chaotic models at horizon 256, zoomed to 1M--2M updates.
  }
  \label{fig:training_scaling_curves}
\end{figure}
\FloatBarrier

\FloatBarrier
\section{Forecasting chaotic dynamics}
\label{sec:chaotic_systems}

Chaotic systems pose a sharp test of forecasting: can a model preserve statistical structure after losing alignment with a particular future?
We compare fine-tuned \chronos{}, PANDA \citep{lai2025panda}, and \method{} on point forecasting and long continuations of 135 DYSTS systems \citep{gilpin2021chaos}.
We evaluate \method{} with ten-step Euler sampling per 256-step block.
\Cref{app:chaos} gives its training recipe, and evaluation details.

\subsection{Forecast accuracy and dynamical fidelity}
Using sMAPE as in PANDA \citep{lai2025panda}, we evaluate forecasts at horizons 64, 128, 256, and 512.
Four complementary diagnostics assess continued paths: Wasserstein-1 distance to an empirical approximation of the invariant distribution, correlation-dimension error, squared Hellinger distance between normalized power spectral densities (PSD), and segment analog score.
Together, they assess state occupancy, geometric dimension, frequency content, and similarity of short multivariate segments to reference dynamics.
Held-out truth provides a finite-trajectory reference.

\begin{table}[!htb]
  \centering
  \setlength{\abovecaptionskip}{0pt}
  \setlength{\belowcaptionskip}{5pt}
  \small
  \setlength{\tabcolsep}{2.5pt}
  \renewcommand{\arraystretch}{1.0}
  \caption{Point accuracy and dynamical fidelity (lower is better; best in bold).
  }
  \label{tab:chaos_comparison}
  \begin{tabular*}{\linewidth}{@{\extracolsep{\fill}}lrrrrrrrr@{}}
    \toprule
    & \multicolumn{4}{c}{sMAPE} & \multicolumn{4}{c}{Dynamics} \\
    \cmidrule(lr){2-5}\cmidrule(l){6-9}
    Model & $H=64$ & $H=128$ & $H=256$ & $H=512$ & $W_1$ & PSD $H^2$ & CD err. & Seg.\ analog \\
    \midrule
    Chronos-1 FT & 24.80 & 31.86 & 39.69 & 47.24 & 0.411 & 0.148 & 0.750 & 16.40 \\
    Chronos-2 FT & 21.07 & 26.78 & 33.28 & 40.07 & 9.686 & 0.269 & 0.569 & 11.12 \\
    PANDA & 19.09 & 24.42 & 31.94 & 41.22 & 0.588 & 0.195 & 0.537 & 10.03 \\
    \method{} (Ours) & \textbf{16.45} & \textbf{21.52} & \textbf{27.43} & \textbf{34.50} & \textbf{0.251} & \textbf{0.061} & \textbf{0.273} & \textbf{4.80} \\
    \midrule
    Held-out truth & --- & --- & --- & --- & 0.047 & 0.014 & 0.055 & 1.00 \\
    \bottomrule
  \end{tabular*}
\end{table}

It achieves substantially better dynamical fidelity over 4,096-step rollouts than all evaluated baselines (\Cref{tab:chaos_comparison}).
Relative to PANDA, it reduces the four long-term diagnostic scores by 49--69\%.
These gains accompany the lowest pointwise error at every reported horizon and persist when \method{} is scaled down to approximately PANDA's 72M parameter count (\Cref{app:chaos_size_sampler}).

Chronos-1 \citep{ansari2024chronos}, an autoregressive forecasting model, has been shown to mimic local chaotic dynamics \citep{yu2026understanding}.
Chronos-1 achieves lower $W_1$ and PSD $H^2$ than the quantile model Chronos-2, with higher pointwise, correlation-dimension, and segment analog errors.
These results highlight the distinct strengths and weaknesses of traditional autoregressive and quantile forecasting models.
\method{} improves across every reported metric, combining stronger point accuracy with greater dynamical fidelity.

Distillation increases point-forecast error relative to the ten-step sampler.
We hypothesize that sensitivity to perturbations amplifies small distillation errors during autoregressive forecasts.
With one or two distilled steps, \method{} still outperforms PANDA on point accuracy and all four dynamical diagnostics.
\Cref{app:chaos_size_sampler} examines this trade-off across model sizes and sampling budgets.

\Needspace{7\baselineskip}
\subsection{What sampling preserves}

\begin{figure}[!htb]
  \centering
  \includegraphics[width=\linewidth,trim=0 3bp 0 4bp,clip]{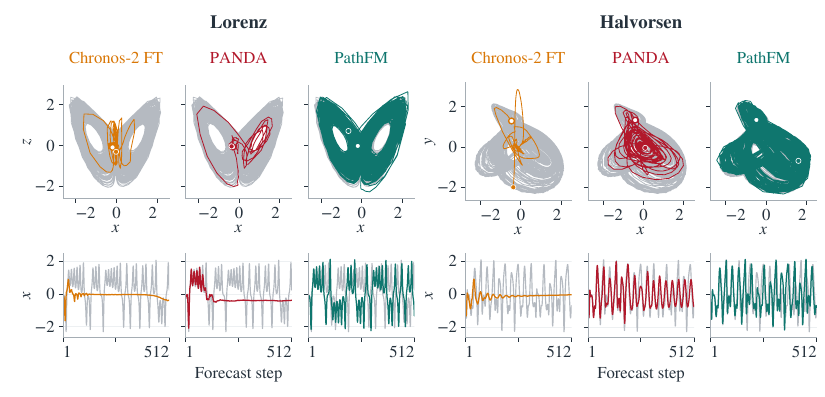}
  \caption{Lorenz and Halvorsen continuations; \method{} uses one sampled path.
  Each column shows a full 4,096-step portrait above its first 512 $x$-coordinate steps.
  Gray curves show the observed continuation; open and filled circles mark the first and last forecast states.
  }
  \label{fig:chaos_comparison}
\end{figure}

\Cref{fig:chaos_comparison} pairs long continuations with their early time traces for Lorenz and Halvorsen systems.
The sampled Lorenz path retains a two-lobe portrait and sustained $x$ variability, whereas the plotted baselines collapse within 256 steps.
For Halvorsen, \method{}'s sample retains the broad three-loop form, while PANDA follows a smaller recurrent orbit.
\Cref{app:chaos} provides further diagnostics.

\FloatBarrier
\Needspace{12\baselineskip}
\section{Discussion and conclusion}
\label{sec:discussion}
\label{sec:conclusion}

\method{} combines flow matching, distillation, shared computation, and direct score refinement to generate complete future trajectories efficiently.
Its strongest gains concern the quality of complete trajectories and the downstream predictions they support.
The dynamical study shows a unique benefit: independently continued paths preserve variability that point forecasts can lose.

Several limitations remain.
\method{} does not yet support conditioning on modalities beyond time series, such as text.
For chaotic systems, all four dynamical diagnostics remain well above the held-out-truth reference (\Cref{tab:chaos_comparison}).
Accurately recovering the invariant distribution remains an open challenge.
A key direction for future work is to continue scaling the backbone and training data, pushing the frontier of generative time series forecasting.

\endgroup

\subsection*{Acknowledgments}
We thank William Gilpin, Jeffrey Lai, and Anthony Bao for their insightful discussions and generous help.
This work was conducted at Amazon AWS.

\subsection*{AI use statement}
We used generative AI tools to support research execution, including clean and reformat dataset, implementation and analysis of experimental results.
We also used these tools for literature retrieval and discovery, drafting sections of the paper, and polishing the writing for clarity and readability.
We have reviewed all AI-assisted work, including code, experimental analyses, literature references, and manuscript text.
We take responsibility for the final content of this work, including text, claims, and artifacts produced with the aid of generative AI.

\subsection*{Reproducibility statement}
\Cref{sec:method} describes the model and learning objectives, and \Cref{app:model_training} provides the architecture and training recipe.
\Cref{sec:fps_bench,app:evaluation} specify the trajectory benchmark, score estimators, and aggregation rules.
\Cref{app:chaos} describes the chaotic-system experiments and their evaluation metrics.
We will release the code and model weights to support reproducibility.

\bibliography{references}
\bibliographystyle{template/iclr2027_conference}

\clearpage
\appendix
\raggedbottom
\clearpage

\section{Related Work}
\label{sec:related-work}

\paragraph{Marginal forecasting.}
Chronos-2, TimesFM-3, and Toto 2.0 provide direct marginal quantiles while sharing information across variables \citep{ansari2025chronos2,timesfm3,khwaja2026toto2}.
Toto 1.0 instead predicts marginal Student-$t$ mixtures and samples forecasts autoregressively across output patches \citep{cohen2025toto}.
\citet{perezdiaz2025form} distinguish forecast products by their operational uses and show that marginals do not identify path-dependent event probabilities.
\method{} represents the joint forecast through complete multivariate paths, so the same samples support marginal summaries and distributions of trajectory-dependent quantities.

\paragraph{Generative trajectory forecasting.}
DeepAR samples temporally dependent futures autoregressively \citep{salinas2020deepar}, and TACTiS learns multivariate predictive distributions with attentional copulas \citep{drouin2022tactis}.
TimeGrad introduces autoregressive diffusion for multivariate forecasting \citep{rasul2021timegrad}.
Flow matching learns generative vector fields through conditional probability paths \citep{lipman2023flow,huang2026recursiveflowmatching}; TSFlow incorporates Gaussian-process priors \citep{kollovieh2024tsflow}, and Sundial uses flow matching for foundation-model pretraining \citep{liu2025sundial}.
\method{} shares history computation across paths and uses distillation to support fast ensembles and direct forecast-score training.

\paragraph{Distillation and direct score optimization.}
Flow map methods learn finite-time transport and support both teacher distillation and self-distillation \citep{boffi2024flowmaps,boffi2025selfdistillation}.
Parallel Decoding Distillation (PDD) predicts multiple integration updates per network evaluation \citep{pdd2026}.
DRaFT optimizes differentiable rewards by backpropagating through diffusion sampling \citep{clark2024draft}.
Rewarded Moment Matching Distillation (RMMD) combines reward optimization with moment-matching supervision and applies unbiased sample CRPS to GenCast \citep{rmmd2026}.
Its reward gradient passes through a denoising step applied to a re-noised student sample.

\method{} backpropagates fair CRPS through the complete source-to-forecast map of one distilled evaluation.
The same optimization updates continue matching the frozen teacher's joint velocity field while fitting the student's marginals to observations.
Distillation thus supplies both a short path for score gradients and continued supervision from a joint generative teacher.
\Cref{sec:method_posttraining,tab:architecture_controls} give the objective and post-training comparisons.

\paragraph{Forecast evaluation.}
Proper scoring rules underpin forecast comparison \citep{gneiting2007proper}.
Fair scores address finite-ensemble effects \citep{ferro2014fair}; variogram and transformation-based scores expose complementary multivariate properties \citep{scheuerer2015variogram,pic2025scores}.
fps-bench combines these principles on fev-bench tasks \citep{fevbench} to assess temporal dependence and trajectory-dependent quantities alongside marginal accuracy.
DYSTS supplies interpretable chaotic systems \citep{gilpin2021chaos}, while PANDA studies pretraining across chaotic dynamics \citep{lai2025panda}.
Together, these evaluations test the value of complete forecast paths through downstream uncertainty, short-term tracking, and long-run dynamical fidelity.

\subsection{Concurrent Work}
\label{app:concurrent-work}

Time series foundation models are advancing rapidly.
t0-gamma alternates temporal and variate attention to incorporate historical and known-future covariates~\citep{meyer2026t0}.
It averages weights from five fine-tuned checkpoints into one 256M-parameter model~\citep{forecastingcompany2026t0gamma}.
QiYao-M fuses numerical patches with text and image embeddings~\citep{cheng2026qiyaom}, but treats each target channel independently.
EXAONE 1.0 specializes at financial forecasting and mixes time with causal convolutions and variates with groupwise pooling MLPs~\citep{lgai2026exaonefinance}.

All three models predict marginal quantiles, leaving dependence across future times and variables unspecified.
\method{} retains its advantage in directly generating joint multivariate trajectories, combining learned temporal and cross-variable dependence with fast sampling and marginal calibration.

\FloatBarrier
\clearpage
\section{Model and training}
\label{app:model_training}
\suppressfloats[t]

This section specifies the backbone and the training recipe for 126M and 300M models, then examines distillation, calibration, and the architecture controls that support the design.
It closes with model-size comparisons of forecast accuracy and inference cost, including results on FEV, FPS, TIME, and GIFT-Eval.
The flow model has 125.1M parameters, and its distilled version has 126.3M including the PDD heads.

\subsection{Backbone and input}
\label{app:representation}

\paragraph{Backbone configuration.}
The settings below describe the general-domain forecasting model; \Cref{app:chaos} gives the separate dynamical configuration.
The backbone uses velocity prediction, ReLU activation, rotary temporal positions, and dynamic adaptive RMS normalization.
The latter applies $(1+\alpha)\odot\operatorname{RMSNorm}(h)+\beta$, with scale and shift obtained from a shared time projection plus block-specific offsets.
Modulation and velocity readouts start at zero; the learned state-entry gate starts at $0.1$.
Temporal and channel attention use query--key normalization.
The Gaussian source scale is fixed at $\sigma=0.75$.
Both model sizes use width 768, 12 attention heads, and feed-forward width 3,072.
The 126M and 300M backbones have 12 and 30 blocks, with two and ten flow-free entry blocks, respectively.

\paragraph{Input representation.}
Following the input representation of Chronos-2~\citep{ansari2025chronos2}, each channel uses context-derived location $\mu_c$ and positive scale $s_c$:
\begin{equation}
  \bar Y_{\ell c}=\operatorname{asinh}\!\left(\frac{Y_{\ell c}-\mu_c}{s_c}\right),
  \qquad
  Y_{\ell c}=\mu_c+s_c\sinh(\bar Y_{\ell c}).
  \label{eq:method_normalization}
\end{equation}
The scale is the context root-mean-square deviation, floored at $10^{-4}$.
Let binary masks $V$ and $K$ mark valid requested coordinates and known future coordinates, respectively, and let $A$ contain normalized known values, zero elsewhere.
Only $M=V\odot(1-K)$ is transported.
We write
\begin{equation}
  \Pi_M(X)=M\odot X+K\odot A,
  \qquad
  \|B\|_M^2:=
  \frac{\sum_{\ell,c}M_{\ell c}B_{\ell c}^2}
       {\max(1,\sum_{\ell,c}M_{\ell c})},
  \label{eq:method_masks}
\end{equation}
Here $\odot$ denotes elementwise multiplication; batch axes are included when present.
The supervision mask excludes missing targets, known covariates, and final-patch padding.

\paragraph{Flow-time sampling.}
Each training row uses one scalar flow time, shared across all channels and forecast steps.
For each row and draw, we sample $t=\operatorname{sigmoid}(-0.8+0.8\epsilon)$, $\epsilon\sim\mathcal N(0,1)$.
The interpolated state and velocity target follow \Cref{eq:method_fm}, with this same $t$ applied to the entire future.
Four independent source/time draws share each history, with their losses averaged and normalized by the batch's transported-coordinate count.
At inference, all channels advance at a common flow time.

\subsection{Training recipe}
\label{app:training_recipe}

Both 126M and 300M models follow the four-stage schedule in \Cref{tab:training_schedule}.
Transformer weights are initialized afresh while retaining the released \chronos{} input embeddings, which remain trainable.

\begin{table}[H]
  \centering
  \small
  \setlength{\tabcolsep}{6pt}
  \caption{General-corpus training schedule for both model sizes.
  Warmup is linear and included in the update budget.
  Flow matching and PDD use constant rates after warmup; WSD decays linearly to zero.
  The final stage decays linearly to zero for 126M and uses a constant rate for 300M.
  EMA starts at the end of warmup and is preserved across continuation stages.
  Each complete recipe contains 2.3M optimizer updates.}
  \label{tab:training_schedule}
  \begin{tabular}{lrrrr}
    \toprule
    Stage & Updates & Peak learning rate & Warmup & EMA decay \\
    \midrule
    Flow matching & 2,000,000 & $5\times10^{-5}$ & 2,000 & 0.9999 \\
    Terminal decay (WSD) & 100,000 & $5\times10^{-5}$ & --- & 0.9999 \\
    PDD & 100,000 & $3\times10^{-5}$ & 500 & 0.999 \\
    PDD + fair CRPS & 100,000 & $10^{-5}$ & --- & 0.999 \\
    \bottomrule
  \end{tabular}
\end{table}

\paragraph{Optimization and stage transitions.}
The recipe uses AdamW with $(\beta_1,\beta_2)=(0.9,0.99)$ and global gradient-norm clipping at 1.
Both models use weight decay 0.01 during flow matching and WSD.
An optimizer update contains 1,024 channel rows.
Terminal decay continues the flow model with its optimizer and EMA state.
PDD initializes the student from the terminal teacher's EMA weights, freezes that teacher, and starts a fresh optimizer.
Calibration resumes the PDD student, optimizer, and EMA.

\paragraph{Data and task sampling.}
Training uses a 83M-row dataset of real-world and synthetic time series, including both univariate and multivariate data.
Training samples contexts up to 8,192 time steps and forecast horizons up to 480 time steps.
During training, we randomly sample context and forecast lengths within these limits.

\paragraph{Post-training objectives.}
PDD trains the backbone and twenty interval heads on a uniform time grid, using the block-size menu $\{1,2,5,10,20\}$, on-policy states, Euler teacher labels, and tiled residual heads.
Calibration retains path-wise PDD and adds fair CRPS with coefficient one on each loss, globally normalized over scored coordinates.
The CRPS term uses eight iid paths, one parallel evaluation, and the same Gaussian source scale as flow training and inference.

Main benchmark results use the final EMA checkpoint after all four stages.
The architecture, distillation, and refinement controls below retain their stated training budgets.

\FloatBarrier
\subsection{Distillation and evaluation budget}
\label{app:pdd_nfe}

\begin{figure}[!htbp]
  \centering
  \includegraphics[width=\linewidth]{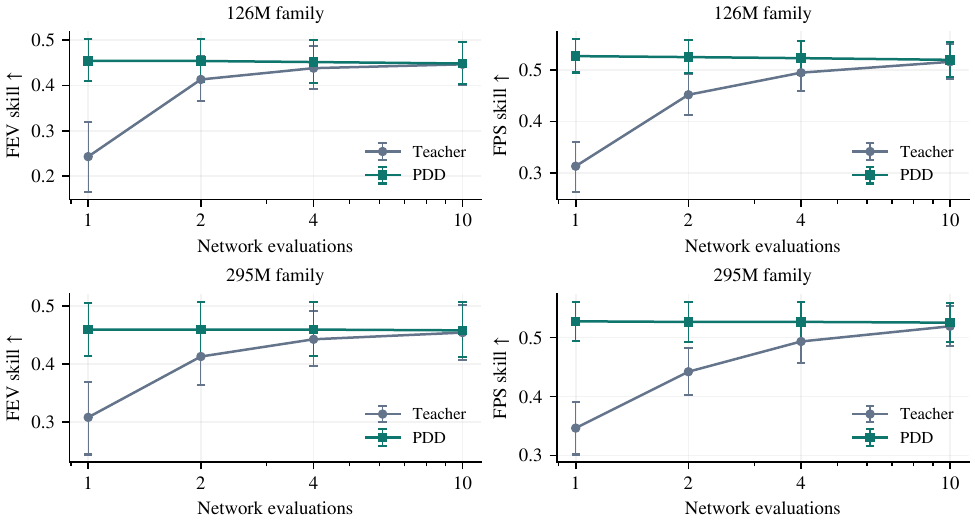}
  \caption{Teacher--student accuracy at different evaluation budgets.
  Rows show teacher--student pairs from the 126M and 295M families; columns show FEV and six-component FPS skill, with higher values preferred.
  NFE uses a logarithmic horizontal axis.
  Thin vertical bars show pointwise 95\% task-bootstrap intervals from 2,000 resamples.}
  \label{fig:pdd_teacher_nfe}
\end{figure}

\Cref{fig:pdd_teacher_nfe} reports a teacher--student inference budget comparison in both model sizes.
Each pair is evaluated at 1, 2, 4, and 10 sequential network evaluations (NFE).

At one evaluation, PDD avoids the teacher's sharp loss of accuracy and has higher skill scores than the ten-evaluation teacher in both pairs.
Additional evaluations give no consistent improvement.
These results show that the student preserves the teacher's forecasting accuracy at a much lower budget.
\FloatBarrier
\clearpage
\subsection{Architecture and training ablations}
\label{app:ablation}

\Cref{tab:architecture_controls} follows the cumulative design at the controlled development budget, from flow matching (R0--R5) through distillation, fair-CRPS refinement, and stratified sampling (R6--R8).
The modulation, initialization, normalization, and regularization bundle in R2 improves both FEV and FPS.
Restricting history attention in R3 enables shared history computation but reduces FEV skill; late entry and four training draws partially recover it.
Distillation then improves both scores while reducing the number of evaluations from ten to one.
R7 adds fair CRPS to the PDD objective, and R8 changes source sampling at the same weights.
Stratification improves empirical FPS from 0.5248 to 0.5329, with a paired 95\% task-bootstrap interval of $[0.0042,0.0131]$ for the skill gain.

\begin{table}[!htbp]
  \centering
  \small
  \setlength{\tabcolsep}{4pt}
  \renewcommand{\arraystretch}{1.12}
  \caption{Cumulative architecture and post-training study.
  Rows R0--R7 use iid sources.
  }
  \label{tab:architecture_controls}
  \begin{tabularx}{\linewidth}{@{}lXrrr@{}}
    \toprule
     & Cumulative change & NFE & FEV $\uparrow$ & FPS $\uparrow$ \\
    \midrule
    R0 & Flow matching with clean target & 10 & 0.4314 & 0.4379 \\
    R1 & + Velocity prediction & 10 & 0.4311 & 0.4216 \\
    R2 & + Time modulation, zero-init head, QK norm, weight decay & 10 & 0.4468 & 0.4960 \\
    R3 & + History-only attention & 10 & 0.4356 & 0.5008 \\
    R4 & + Two flow-free entry blocks & 10 & 0.4381 & 0.5088 \\
    R5 & + Four packed training draws & 10 & 0.4402 & 0.5076 \\
    \midrule
    R6 & + Parallel Decoding Distillation & 1 & 0.4471 & 0.5221 \\
    R7 & + PDD with fair CRPS & 1 & 0.4488 & 0.5248 \\
    \midrule
    R8 & + Stratified sampling & 1 & \textbf{0.4575} & \textbf{0.5329} \\
    \bottomrule
  \end{tabularx}
\end{table}

\FloatBarrier
\subsection{Scaling and additional benchmarks}
\label{app:scaling}
\suppressfloats[t]

We compare the 126M and 300M models on FEV, FPS, GIFT-Eval, and TIME, as well as their cost.
Their training curves are in \Cref{fig:training_scaling_curves}.

\subsubsection{Model size and forecast accuracy}
\label{app:benchmark_transfer}
The 126M and 300M families contain 126.3M and 296.3M learned parameters, respectively, including their distillation heads.
Their training recipes are given in \Cref{app:training_recipe}.

\begin{figure}[!htbp]
  \centering
  \includegraphics[width=\linewidth]{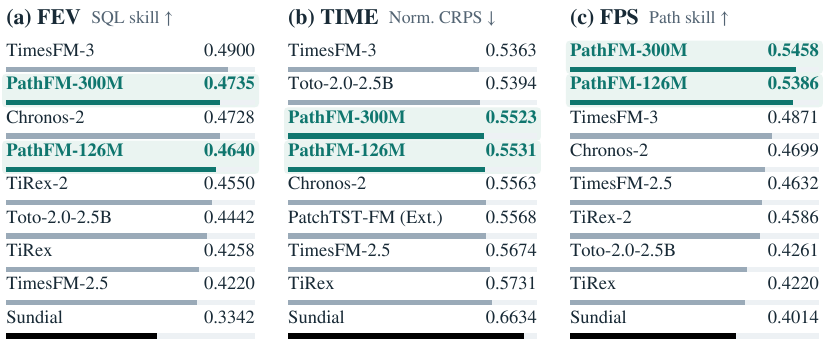}
  \caption{Scaling \method{} from the 126M to the 300M family.
  Panels rank nine models on FEV SQL skill (100 tasks), TIME normalized CRPS (98 tasks), and FPS path skill (94 tasks).
  Native samplers \method{} and Sundial are teal and black.
  FPS selects each quantile model's better independent or comonotone completion.}
  \label{fig:pathfm_scaling}
\end{figure}

The 300M \method{} variant has higher FEV and FPS skill and lower GIFT-Eval and TIME CRPS than the 126M model (\Cref{fig:pathfm_scaling,tab:benchmark_transfer}).
\method{}-300M ranks second on FEV, third on TIME, and first on FPS among the baselines.

\Cref{tab:benchmark_transfer} adds GIFT-Eval \citep{gift_eval} to the benchmark suite.
GIFT uses all 97 configurations without series or window caps and with cross-learning disabled.
Against the external baselines, our 126M model ranks fourth on GIFT-Eval CRPS and the 300M model ranks third.

\begin{table}[!htbp]
  \centering
  \small
  \setlength{\tabcolsep}{6pt}
  \colorlet{benchblue}{backboneblue!30}
  \newcommand{\benchcell}[2]{\cellcolor{benchblue!#1}#2}
  \caption{Forecasting accuracy across four benchmarks.
  Darker blue indicates better performance on a linear scale within each column; bold marks the best value.
  Dashes indicate unavailable results.}
  \label{tab:benchmark_transfer}
  \begin{tabular}{lrrrrrr}
    \toprule
    & FEV & FPS & \multicolumn{2}{c}{GIFT-Eval} & \multicolumn{2}{c}{TIME} \\
    \cmidrule(lr){2-2}\cmidrule(lr){3-3}\cmidrule(lr){4-5}\cmidrule(lr){6-7}
    Model & Skill $\uparrow$ & Skill $\uparrow$ & MASE $\downarrow$ & CRPS $\downarrow$ & MASE $\downarrow$ & CRPS $\downarrow$ \\
    \midrule
    TimesFM-3 & \benchcell{55.0}{\textbf{0.4900}} & \benchcell{34.3}{0.4871} & --- & --- & \benchcell{55.0}{\textbf{0.6398}} & \benchcell{55.0}{\textbf{0.5363}} \\
    Chronos-2 & \benchcell{49.4}{0.4728} & \benchcell{28.2}{0.4699} & \benchcell{52.9}{0.6978} & \benchcell{49.2}{0.4854} & \benchcell{45.4}{0.6620} & \benchcell{47.0}{0.5563} \\
    TiRex-2 & \benchcell{43.5}{0.4550} & \benchcell{24.2}{0.4586} & \benchcell{53.4}{0.6973} & \benchcell{53.6}{0.4781} & --- & --- \\
    Toto-2.0-2.5B & \benchcell{40.0}{0.4442} & \benchcell{12.7}{0.4261} & \benchcell{55.0}{\textbf{0.6956}} & \benchcell{55.0}{\textbf{0.4759}} & \benchcell{54.1}{0.6419} & \benchcell{53.8}{0.5394} \\
    TiRex & \benchcell{34.0}{0.4258} & \benchcell{11.3}{0.4220} & \benchcell{36.1}{0.7158} & \benchcell{47.3}{0.4885} & \benchcell{36.3}{0.6831} & \benchcell{40.2}{0.5731} \\
    TimesFM-2.5 & \benchcell{32.7}{0.4220} & \benchcell{25.8}{0.4632} & \benchcell{46.2}{0.7050} & \benchcell{46.2}{0.4903} & \benchcell{42.5}{0.6686} & \benchcell{42.5}{0.5674} \\
    Sundial & \benchcell{4.0}{0.3342} & \benchcell{4.0}{0.4014} & \benchcell{4.0}{0.7502} & \benchcell{4.0}{0.5590} & \benchcell{4.0}{0.7576} & \benchcell{4.0}{0.6634} \\
    \midrule
    \method{}-126M & \benchcell{46.5}{0.4640} & \benchcell{52.5}{0.5386} & \benchcell{44.1}{0.7073} & \benchcell{47.7}{0.4878} & \benchcell{48.0}{0.6560} & \benchcell{48.2}{0.5531} \\
    \method{}-300M & \benchcell{49.6}{0.4735} & \benchcell{55.0}{\textbf{0.5458}} & \benchcell{50.1}{0.7008} & \benchcell{51.4}{0.4818} & \benchcell{48.9}{0.6538} & \benchcell{48.6}{0.5523} \\
    \bottomrule
  \end{tabular}
\end{table}

\FloatBarrier
\subsubsection{Accuracy and inference cost}
\begin{figure}[!htbp]
  \centering
  \includegraphics[width=\linewidth]{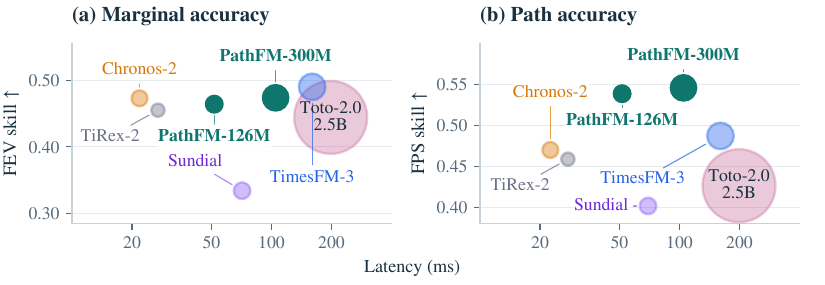}
  \caption{Accuracy, latency, and model size with the 300M \method{} variant.
  Circle area denotes parameter count.
  Latency is the median time to return nine quantiles (a) or forecast paths (b) for one query on an NVIDIA L4 at context 2,048 and horizon 288.
  Accuracy uses the native benchmark settings.}
  \label{fig:pathfm_scaling_latency}
\end{figure}

\Cref{fig:pathfm_scaling_latency} adds the 300M model to the earlier 126M timing comparison in \Cref{fig:model_family_latency}.
Both variants lie on the observed accuracy--latency frontier for path forecasts.
\Cref{app:latency} specifies the matched timing protocol and its limitations.

\FloatBarrier
\clearpage
\section{Sampling and computational efficiency}
\label{app:sampling_efficiency}
\suppressfloats[t]

Ensemble size determines the precision of forecast summaries, while the sampler and its execution determine their cost.
We first describe sampling error and ensemble construction, then report controlled latency and memory comparisons.

\subsection{Finite-ensemble approximation}
\label{app:mc_limit}

A finite ensemble approximates a model's predictive distribution through sampled paths.
Forecast summaries and scores can vary with the returned ensemble due to Monte Carlo approximation.
We first separate approximation error from model error, then describe ensemble construction; \Cref{app:score_estimation} defines the scores of the empirical forecast.

\subsubsection{Forecast quantities and model error}

Condition on the available forecasting information $\mathcal I$ and on the fitted model.
For $m\geq1$ iid paths from $P_\theta(\cdot\mid\mathcal I)$ and a fixed functional $g$ of finite variance, write $\mu_\theta=\mathbb E_{P_\theta}[g(X)\mid\mathcal I]$.
The sample average is unbiased for $\mu_\theta$, and independence gives
\begin{equation}
 \widehat\mu_g=\frac1m\sum_{i=1}^{m}g(X^{(i)}),\qquad
 \mathbb E_Z[(\widehat\mu_g-\mu_\theta)^2\mid\mathcal I]
 =\frac{\operatorname{Var}_{P_\theta}(g(X)\mid\mathcal I)}{m}.
 \label{eq:mc_limit}
\end{equation}
Thus ordinary iid Monte Carlo standard error scales as $m^{-1/2}$.
This error remains for nondegenerate queries even when $P_\theta$ is the true conditional law.
Direct quantile outputs avoid estimating their reported quantiles from finitely many paths, but retain model error.
Efficient ensemble generation therefore matters for both inference cost and the accuracy of downstream estimates.

Let the finite value $\mu_\star$ be the corresponding expectation under the true conditional law.
The sample average's squared error relative to this target decomposes as
\begin{equation}
 \mathbb E_Z[(\widehat\mu_g-\mu_\star)^2\mid\mathcal I]
 =(\mu_\theta-\mu_\star)^2
 +\frac{\operatorname{Var}_{P_\theta}(g(X)\mid\mathcal I)}{m}.
 \label{eq:mc_decomposition}
\end{equation}
The cross term vanishes because $\mathbb E_Z[\widehat\mu_g-\mu_\theta\mid\mathcal I]=0$.
The first term is error in the model's conditional functional expectation; the second is Monte Carlo approximation error.
This expression concerns estimation of a conditional expectation, not prediction error against one random future outcome, which also contains outcome variability.

For an event $A$ with model probability $p_\theta=P_\theta(A\mid\mathcal I)$, taking $g(X)=\mathbf 1\{X\in A\}$ gives sampling variance $p_\theta(1-p_\theta)/m$.
For $0<p_\theta<1$, the relative standard error is $\sqrt{(1-p_\theta)/(m p_\theta)}$, making rare-event probability estimates demanding at small ensemble sizes.
Increasing $m$ reduces this sampling term but does not remove model error.
Using conditional marginals or analytical integration, when available for a particular query, need not incur the same sampling error.
Dependent or stratified draws require a different variance analysis.
The $m^{-1/2}$ rate describes the approximation error of the iid sample average at a finite sample budget.

\subsubsection{Ensemble construction and stratification}
\label{app:ensemble_sampling}

At inference, we approximate the forecast distribution with an ensemble of $m$ complete future trajectories, from which we estimate quantiles and other forecast summaries.
Ensemble size controls the number of paths, independently of the one- or two-evaluation budget per path.
With few draws, these summaries can be sensitive to sampling variability.
\Cref{app:marginal_coverage} measures the effect of sample count on marginal interval coverage across all 100 fev-bench test tasks.
For more even source coverage at fixed $m$, each coordinate uses the midpoints of $m$ equiprobable Gaussian bins, independently permuted across paths:
\begin{equation}
  u_r=\frac{r-\tfrac12}{m},\qquad
  Z^j_{\ell c}=\sigma\Phi^{-1}(u_{\pi_{\ell c}(j)}),
  \quad r,j\in\{1,\ldots,m\},
  \label{eq:method_stratification}
\end{equation}
Here $\Phi$ is the standard-normal CDF and each $\pi_{\ell c}$ is an independent permutation.
This assigns every rank once per coordinate without tying a path's ranks across times or channels.

\FloatBarrier
\subsection{Marginal coverage and ensemble size}
\label{app:calibration_controls}
\label{app:marginal_coverage}

\begin{figure}[!htbp]
  \centering
  \includegraphics[width=\linewidth,trim=0 10bp 0 0bp,clip]{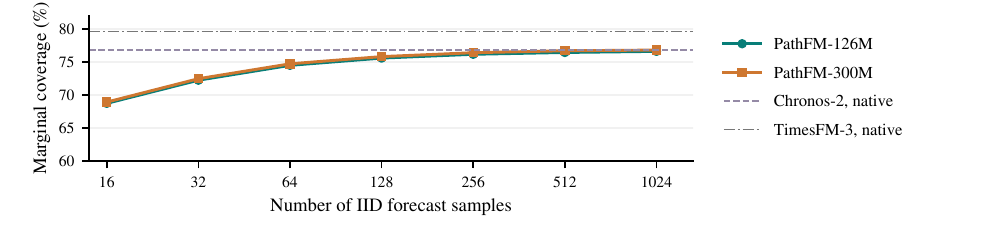}
  \caption{Marginal 80\% interval coverage versus the number of forecast samples on fev-bench.
  Both model sizes use iid sampling.
  Horizontal references use the native 10th and 90th percentile outputs of \chronos{} and TimesFM-3.}
  \label{fig:marginal_coverage_samples}
\end{figure}

Larger ensembles improve marginal interval coverage at fixed model weights (\Cref{fig:marginal_coverage_samples}).
Across all 100 fev-bench tasks, increasing the ensemble from 32 to 1,024 iid samples raises 80\% interval coverage by 4 percentage points.
The gains diminish as ensemble size increases.

\FloatBarrier

\subsection{Inference latency}
\label{app:latency}

\Cref{tab:latency_primary} compares complete serving calls on an NVIDIA L4 with context 2,048, horizon 288, and batches of one or sixteen histories.
We measure end-to-end time, with CUDA synchronization around each call.
We report medians over 35 measured calls in seven blocks, following warmup and with randomized execution order.
Models retain their native behavior.
The marginal interface returns nine quantiles.
The path interface returns forecast trajectories, using independent completion for quantile outputs.

\begin{table}[!htbp]
  \centering
  \caption{End-to-end latency on an NVIDIA L4 at context 2,048 and horizon 288.}
  \label{tab:latency_primary}
  \small
  \setlength{\tabcolsep}{4pt}
\begin{tabularx}{\linewidth}{@{}Xrrrrr@{}}
  \toprule
  & & \multicolumn{2}{c}{Nine quantiles (ms)} & \multicolumn{2}{c}{Paths (ms)} \\
  \cmidrule(lr){3-4}\cmidrule(lr){5-6}
  Model & Params.\ (M) & Batch 1 & Batch 16 & Batch 1 & Batch 16 \\
  \midrule
  \rowcolor{gray!20}
  \method{}, one evaluation & 126.3 & 51.7 & 215.3 & 51.6 & 215.2 \\
  \method{}, two evaluations & 126.3 & 96.4 & 403.2 & 96.4 & 403.3 \\
  \chronos{} & 119.5 & 21.8 & 66.6 & 22.5 & 72.9 \\
  TimesFM-3 & 330.7 & 160.0 & 508.2 & 160.6 & 515.1 \\
  Toto-2.0-2.5B & 2,454.3 & 198.3 & 1,482.6 & 199.0 & 1,489.5 \\
  TiRex & 35.3 & 170.4 & 180.9 & 170.9 & 187.1 \\
  TiRex-2 & 82.5 & 27.0 & 44.0 & 27.6 & 50.5 \\
  Sundial & 128.3 & 71.2 & 1,118.3 & 69.6 & 1,085.1 \\
  \bottomrule
\end{tabularx}
\end{table}

\method{} returns an ensemble in 51.6\,ms for one history and 215.2\,ms for sixteen.
Batching sixteen histories reduces the average time per history to 13.5\,ms.
At both batch sizes, \method{} with one evaluation is faster than Sundial and TimesFM-3.
\chronos{} and TiRex-2 offer lower latency, while \method{} achieves higher FPS skill (\Cref{fig:model_family_latency}).

\subsubsection{Inference speedup breakdown}

\begin{figure}[!htbp]
  \centering
  \includegraphics[width=0.6\linewidth]{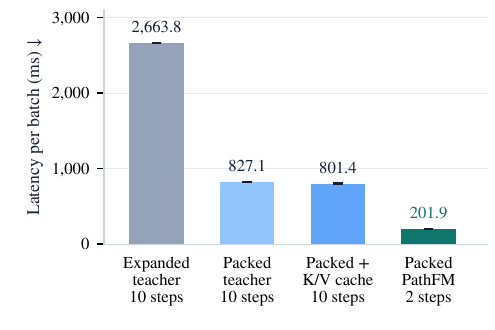}
  \caption{Inference speedup breakdown on an A100 at batch 16, context 2,048, and horizon 480.
  The teacher uses ten Euler evaluations; the student uses two.
  All configurations use float32 weights with bfloat16 autocast.}
  \label{fig:sampling_tradeoffs}
\end{figure}

\Cref{fig:sampling_tradeoffs} isolates execution choices in a separate study on A100.
Packing paths reduces teacher latency from 2,663.8 to 827.1\,ms per batch; the packed two-evaluation student takes 201.9\,ms.
Adding deeper history K/V caching changes teacher latency only to 801.4\,ms and is disabled for the student.
The combined student speedup therefore comes from packing and distillation.
Packing also reduces peak PyTorch allocated memory from 3.22 to 1.58\,GiB.

\FloatBarrier
\section{Evaluation metrics}
\label{app:evaluation}
\suppressfloats[t]

We specify the empirical forecast scores and aggregation rules used in the forecasting results.
The chaotic-system protocols and diagnostics are given separately in \Cref{app:chaos}.

\subsection{Score construction}
\label{app:metrics}

\subsubsection{Empirical forecast scores}
\label{app:score_estimation}

For a fixed outcome $y$ and predictive law $P$, CRPS and energy score have the common form
\begin{equation}
 S_\rho(P,y)=\mathbb E\rho(X,y)-\tfrac12\mathbb E\rho(X,X'),
 \qquad X,X'\stackrel{\mathrm{iid}}{\sim}P,
\end{equation}
using absolute and Euclidean distance, respectively \citep{gneiting2007proper}.
We evaluate the returned empirical forecast $\widehat P_m=\frac{1}{m}\sum_{i=1}^{m}\delta_{x^{(i)}}$.
Substituting this distribution gives
\begin{equation}
 S_\rho(\widehat P_m,y)
 =\frac1m\sum_{i=1}^{m}\rho(x^{(i)},y)
 -\frac{1}{2m^2}\sum_{i,j=1}^{m}\rho(x^{(i)},x^{(j)}).
 \label{eq:empirical_score}
\end{equation}
This is the exact score of the finite forecast distribution and requires no independence assumption on its members.
We use \Cref{eq:empirical_score} for every CRPS and energy component.
The variogram component evaluates the moments of the same empirical forecast (\Cref{eq:variogram_score}).

\subsubsection{Temporal path scores}

For item $n$ and channel $c$, define the seasonal-error scale from history $h$ and seasonal period $p$:
\begin{equation}
 a_{nc}=\frac{1}{|\mathcal H_{nc}|}
 \sum_{t\in\mathcal H_{nc}}|h_{ntc}-h_{n,t-p,c}|,
 \qquad
 \mathcal H_{nc}=\{t:h_{ntc},h_{n,t-p,c}\text{ observed}\}.
 \label{eq:seasonal_scale}
\end{equation}
Except for first passage, the temporal scores use paths divided by finite positive $a_{nc}$.
\Cref{tab:score_objects} specifies their distances in \Cref{eq:empirical_score}; the variogram uses its separate formula below.

\begin{table}[!htbp]
 \centering
 \small
 \setlength{\tabcolsep}{5pt}
 \caption{The definitions of the six temporal FPS components.}
 \label{tab:score_objects}
 \begin{tabularx}{\linewidth}{@{}lX@{}}
  \toprule
  Score & Distance or transformation \\
  \midrule
  Energy (ES) & $\rho(x,y)=\|x-y\|_2$ \\
  Differenced ES (DES) & $\rho(x,y)=\|D(x)-D(y)\|_2$, where $D(x)=(x_1,x_2-x_1,\ldots,x_L-x_{L-1})$ \\
  Variogram (VS) & Squared errors in expected temporal contrasts, \Cref{eq:variogram_score} \\
  Sum CRPS & $\rho(x,y)=|\sum_t x_t-\sum_t y_t|$ \\
  Max CRPS & $\rho(x,y)=|\max_t x_t-\max_t y_t|$ \\
  First-passage CRPS & $\rho(u,v)=|T_\tau(u)-T_\tau(v)|$ in original units, \Cref{eq:first_passage_crps} \\
  \bottomrule
 \end{tabularx}
\end{table}

Level, sum, and maximum scores use the same observed-coordinate mask for every forecast path.
DES retains the initial value when observed and increments only when both adjacent observations are available.

\paragraph{Variogram.}
For the nonempty set of observed time pairs $\mathcal P=\{(s,t):s<t,\ y_s,y_t\text{ observed}\}$:
\begin{equation}
 \widehat{\mathrm{VS}}_{1/2}
 =\frac{1}{|\mathcal P|}\sum_{(s,t)\in\mathcal P}
 \left(
 |y_s-y_t|^{1/2}
 -\frac1m\sum_{i=1}^{m}|x_s^{(i)}-x_t^{(i)}|^{1/2}
 \right)^2.
 \label{eq:variogram_score}
\end{equation}
All pairs receive equal weight \citep{scheuerer2015variogram}.

\paragraph{First passage.}
The threshold $\tau$ is the maximum of the last available context segment of length at most $L$, in original units.
For a path $u$,
\begin{equation}
 T_\tau(u)=\min\bigl(\{t\in\{1,\ldots,L\}:u_t>\tau\}\cup\{L+1\}\bigr).
 \label{eq:first_passage_crps}
\end{equation}
This score requires a fully observed future and finite threshold, uses forecast steps as units, and treats no crossing as $L+1$.
It therefore evaluates a censored event time.
The sum, maximum, and first-passage scores identify their respective functional distributions rather than the full path law.

\subsubsection{Cross-variable variogram score}
\label{app:joint_metrics}

The cross-variable variogram compares channel pairs at the same forecast time.
Channels are standardized by their mean and standard deviation over at most 8,192 context observations; insufficient history or negligible variance excludes a channel.
Using shared observed-coordinate masks, the estimator in \Cref{eq:variogram_score} averages over valid times and channel pairs, items, origins, and finally tasks with an arithmetic mean.
We report this absolute score.

\subsection{Aggregation rules}

\paragraph{Marginal benchmarks.}
We retain the official within-task reductions for FEV, GIFT-Eval, and TIME \citep{fevbench,gift_eval,qiao2026time}.
FEV uses nine quantiles at $\{0.1,0.2,\ldots,0.9\}$.
SQL scales errors by $a_{nc}$ before averaging.
GIFT-Eval and TIME report equally weighted geometric means of task or configuration losses relative to Seasonal Naive.

\paragraph{fps-bench.}
FPS compares each model with Seasonal Naive across six path metrics on the same 94 tasks with valid scores for every model and metric.
For each metric, we average valid items within each channel, then channels and forecast origins, to obtain one loss per task.
Let $r_{d,k}$ be the model's loss divided by Seasonal Naive's loss for task $d$ and metric $k$, clipped to $[0.01,100]$.
The aggregate score is
\begin{equation}
 \mathrm{FPS}=1-\operatorname{GM}_{d,k}(r_{d,k}),
 \label{eq:benchmark_skill}
\end{equation}
where $\operatorname{GM}$ denotes the geometric mean with equal weight for each task and metric.
A score of zero matches Seasonal Naive; higher is better.
Individual component skills use the same rule for one metric at a time.

\FloatBarrier
\section{Forecast accuracy and dependence}
\label{app:forecast_quality}
\suppressfloats[t]

These experiments examine which aspects of a forecast improve when the model generates complete paths.
We report temporal component scores, evaluate cross-variable relationships, and test explicit dependence completions of quantile forecasts.
Metric definitions and aggregation rules are collected in \Cref{app:evaluation}.

\subsection{Temporal forecast accuracy}
\label{app:benchmark_details}

\begin{figure}[!htbp]
  \centering
  \includegraphics[width=\linewidth]{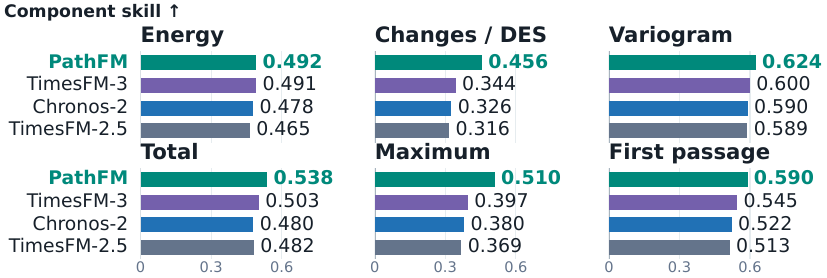}
  \caption{Component skills for \method{} and the three strongest baselines on fps-bench.
  Bars show skill relative to Seasonal Naive; higher is better.}
  \label{fig:fps_component_bars}
\end{figure}

\Cref{fig:fps_component_bars} compares the six FPS components for \method{} and the three strongest baselines (selecting the best-performing completion for each).
\Cref{tab:fps_main} reports the full comparison.
For quantile models, independent completion usually gives better temporal scores than comonotone completion, but the latter can improve sum skill.
\method{} achieves the highest skill scores on all six components, with a 10.6\% average improvement against the strongest baseline, TimesFM-3.

\begin{table}[!htbp]
  \centering
  \small
  \setlength{\tabcolsep}{3pt}
  \caption{Temporal-path skill on the 94 tasks in fps-bench; higher is better.}
  \label{tab:fps_main}
  \begin{tabular}{llrrrrrrr}
    \toprule
    Forecast & Completion & ES & DES & VS & Sum & Max & FPT & FPS skill \\
    \midrule
    TimesFM-3 & Independent & 0.4913 & 0.3437 & 0.5998 & 0.5032 & 0.3966 & 0.5455 & 0.4871 \\
    TimesFM-3 & Comonotone & 0.4757 & 0.3650 & 0.4894 & 0.5049 & 0.3435 & 0.5462 & 0.4590 \\
    Chronos-2 & Independent & 0.4783 & 0.3255 & 0.5901 & 0.4797 & 0.3805 & 0.5224 & 0.4699 \\
    Chronos-2 & Comonotone & 0.4616 & 0.3590 & 0.4818 & 0.4862 & 0.3331 & 0.5148 & 0.4434 \\
    TimesFM-2.5 & Independent & 0.4645 & 0.3164 & 0.5893 & 0.4818 & 0.3687 & 0.5135 & 0.4632 \\
    TiRex-2 & Independent & 0.4608 & 0.3201 & 0.5790 & 0.4663 & 0.3678 & 0.5165 & 0.4586 \\
    Toto-2.0-2.5B & Independent & 0.4355 & 0.2707 & 0.5333 & 0.4339 & 0.3125 & 0.5225 & 0.4261 \\
    TiRex & Independent & 0.4322 & 0.2783 & 0.5419 & 0.4323 & 0.3236 & 0.4828 & 0.4220 \\
    Sundial & Native & 0.3769 & 0.3615 & 0.5072 & 0.4280 & 0.2999 & 0.4141 & 0.4014 \\
    Toto-1.0 & Native & 0.2346 & -0.2072 & 0.3583 & 0.3985 & -0.1313 & 0.4833 & 0.2300 \\
    \rowcolor{gray!20}
    \method{} & Native & \textbf{0.4917} & \textbf{0.4557} & \textbf{0.6244} & \textbf{0.5383} & \textbf{0.5103} & \textbf{0.5896} & \textbf{0.5386} \\
    \bottomrule
  \end{tabular}
\end{table}

\subsection{Cross-variable variogram scores}
\label{app:joint_dependence}
\label{app:cross_variable}

\Cref{tab:cross_variable_vs} compares absolute cross-variable variogram scores on all 35 multi-target fev-bench tasks, retaining all 625 forecast origins.
The score compares pairs of target variables at the same forecast time, using the standardization and reduction in \Cref{app:joint_metrics}.

The Gaussian copula uses shrinkage estimates of normal-score residual-rank correlation and a shared AR(1) temporal factor, fitted on four forecast windows ending before the first test origin.
Six short-history tasks use shorter calibration horizons.
The nine marginal quantiles are interpolated linearly, with constant tails outside probabilities 0.1--0.9.

\paragraph{Task-specific adaptation.}
\label{app:copula_adaptation}
The fitted Gaussian copula estimates dependence parameters from historical forecast--outcome pairs on each benchmark task.
The forecasting weights remain frozen, but the copula introduces task-specific parameter fitting.
We report these adapted references in a separate table group from \method{} and the fixed independent and comonotone completions, which require no additional task-specific parameter fitting at evaluation time.

\begin{table}[htbp]
  \centering
  \small
  \setlength{\tabcolsep}{10pt}
  \caption{Contemporaneous cross-variable variogram scores on 35 multi-target tasks.
  Entries are arithmetic task means; lower is better, with the lowest value in each protocol group bold.
  Fitted Gaussian references use task-specific calibration.}
  \label{tab:cross_variable_vs}
  \begin{tabular}{llr}
    \toprule
    Forecast & Copula & VS $\downarrow$ \\
    \midrule
    \multicolumn{3}{l}{\emph{No task-specific fitting}} \\
    \midrule
    Chronos-2 & Independent & 0.1491 \\
    Chronos-2 & Comonotonic & 0.1648 \\
    TimesFM-3 & Independent & 0.1446 \\
    TimesFM-3 & Comonotonic & 0.1609 \\
    Toto-2.0-2.5B & Independent & 0.1560 \\
    Toto-2.0-2.5B & Comonotonic & 0.1741 \\
    \rowcolor{gray!20}
    \method{} & Native & \textbf{0.1419} \\
    \midrule
    \multicolumn{3}{l}{\emph{Task-calibrated references}} \\
    \midrule
    Chronos-2 & Fitted Gaussian & 0.1469 \\
    TimesFM-3 & Fitted Gaussian & \textbf{0.1420} \\
    Toto-2.0-2.5B & Fitted Gaussian & 0.1533 \\
    \bottomrule
  \end{tabular}
\end{table}
\FloatBarrier
\subsection{Fitted Gaussian temporal copulas}
\label{app:temporal_copulas}

We test whether fitting a Gaussian temporal copula to historical forecast errors improves the path forecasts of frozen quantile models.
Building on the independent and comonotonic completions in \Cref{app:benchmark_details}, we estimate temporal dependence from target-task history while retaining each baseline's marginal quantile forecasts.

\paragraph{Gaussian copula fitting.}
For each baseline and item, four forecast windows ending before the first test origin supply standardized residuals, which we rank-transform to normal scores.
We estimate a lag-one coefficient within these windows, share it across target channels, clip it to $[-0.95,0.95]$, and fix it before testing.
Items with insufficient information use a task-pooled estimate.
All eligible items are retained; 27 tasks use shorter calibration horizons without changing their test windows.
The fitted coefficient defines a stationary latent AR(1) process used to generate temporally dependent ranks for the baseline's interpolated marginal quantiles, with constant tails outside probabilities 0.1--0.9.
This parameterization restricts the fitted temporal dependence to AR(1).

\begin{table}[H]
  \centering
  \small
  \setlength{\tabcolsep}{8pt}
  \caption{Effect of fitting a Gaussian temporal copula on FPS over 94 common fev-bench tasks.
  Each baseline uses the same marginal forecasts across copula completions.
  Fitted Gaussian rows use task-specific calibration.
  Bold marks the best value within each protocol group.}
  \label{tab:temporal_copulas}
  \begin{tabular}{llr}
    \toprule
    Forecast & Copula & FPS $\uparrow$ \\
    \midrule
    \multicolumn{3}{l}{\emph{No task-specific fitting}} \\
    \midrule
    Chronos-2 & Independent & 0.4699 \\
    Chronos-2 & Comonotonic & 0.4434 \\
    TimesFM-3 & Independent & 0.4871 \\
    TimesFM-3 & Comonotonic & 0.4590 \\
    Toto-2.0-2.5B & Independent & 0.4261 \\
    Toto-2.0-2.5B & Comonotonic & 0.4147 \\
    \rowcolor{gray!20}
    \method{} & Native & \textbf{0.5386} \\
    \midrule
    \multicolumn{3}{l}{\emph{Task-calibrated references}} \\
    \midrule
    Chronos-2 & Fitted Gaussian AR(1) & 0.5077 \\
    TimesFM-3 & Fitted Gaussian AR(1) & \textbf{0.5243} \\
    Toto-2.0-2.5B & Fitted Gaussian AR(1) & 0.4765 \\
    \bottomrule
  \end{tabular}
\end{table}

Fitting the Gaussian copula improves FPS from 0.4699 to 0.5077 for Chronos-2, from 0.4871 to 0.5243 for TimesFM-3, and from 0.4261 to 0.4765 for Toto-2.0-2.5B, relative to independent completion.
However, these completions impose a stationary Gaussian AR(1) dependence structure, and \method{} has a higher FPS score without task-specific copula fitting.
\FloatBarrier

\FloatBarrier
\section{Chaotic systems}
\label{app:chaos}
\suppressfloats[t]

We examine how model size, sampling, and fine-tuning affect point accuracy and long-run dynamics.

\subsection{Setup and forecasting protocols}
\label{app:chaos_protocols}

\paragraph{Dataset.}
We train \method{} on the PANDA-72M reproduction corpus \citep{lai2025panda}, \texttt{GilpinLab/dataset-panda72M-repro}, containing 190,006 trajectories of 4,096 steps with 3--10 state coordinates from individual DYSTS systems and skew-product combinations.
Reserving trajectories involving ten parent systems leaves 156,784 training trajectories and 33,222 validation trajectories.
Point evaluation uses the separate \texttt{base40} and \texttt{skew40} test releases.

\paragraph{Models and training.}
The quantitative comparisons use the 75M and 126M \method{} families with native 256-step horizons, trained for 2M flow-matching updates followed by 100k terminal-decay updates.
We train the 75M variant to approximately match PANDA's parameter count (74.2M versus 71.6M).
The 126M backbone has 12 transformer blocks and feedforward width 3,072; the 75M model uses 11 blocks and width 768.
Both initialize their backbones afresh while retaining trainable input-patch embeddings.
The global batch budget is 384 channel rows, with four flow draws per window.
The learning rate is $10^{-4}$ after 500 warmup updates, followed by the 100k linear terminal decay to zero; weight decay is zero and EMA decay is 0.999.
Main results use ten-step Euler sampling, source standard deviation 0.86, and seed 42.

\paragraph{Training-progress evaluation.}
\Cref{fig:training_scaling_curves}c uses the same 2,048 fixed three-channel validation windows for both model sizes, selected randomly from the validation shard.
Each window supplies 512 context steps and 256 forecast steps from a distinct held-out trajectory.
We evaluate EMA weights with 16 stratified samples and ten Euler steps.
Context-standardized sMAPE is averaged over channels and forecast steps, then across the 2,048 trajectories.

\paragraph{Point forecasts.}
We evaluate six 512-context windows from each of 9,347 trajectories using sMAPE, as in PANDA \citep{lai2025panda}.
We average sMAPE within each of the 480 source groups and then report the arithmetic mean across groups.
\method{} feeds back the coordinatewise median of 16 stratified samples per 256-step block, averaging the middle two order statistics.
Chronos-1 uses 16 samples and median feedback every 128 steps, chaining two native 64-step calls within each sample; Chronos-2 and PANDA continue point forecasts in 128-step blocks.

\paragraph{Long continuations.}
For each of the 135 systems, we generate 4,096-step continuations from origins 512 and 4,096 on an unseen trajectory.
\method{} and Chronos-1 independently continue 16 iid paths per origin; Chronos-2 and PANDA continue point forecasts.
For each diagnostic, we take the arithmetic mean over all realizations in a system.
Across the 135 systems, we use a geometric mean for segment analog and arithmetic means for the other diagnostics.
Segment-analog ratios differ by orders of magnitude across systems because their calibration distances can have very different scales; geometric aggregation gives each system equal weight on a log scale.

\subsection{Model size and sampler choice}
\label{app:chaos_size_sampler}

\Cref{tab:chaos_size} compares \method{} with 72M-parameter PANDA.
The smaller \method{} retains lower errors at all four forecasting horizons and on all four dynamical diagnostics, showing that the gains persist at similar model size.
We obtain these gains with our general forecasting architecture and flow-matching objective, without adding architectural components or losses tailored to chaotic systems.
PANDA uses specialized dynamics embeddings with random polynomial and Fourier features motivated by Koopman-operator approximations \citep{lai2025panda}.

\begin{table}[!htbp]
  \centering
  \small
  \setlength{\tabcolsep}{2.5pt}
  \caption{Model size at ten Euler evaluations per block.
  Both \method{} sizes follow the protocol in \Cref{app:chaos_protocols}; lower is better, with the best value in bold.}
  \label{tab:chaos_size}
  \begin{tabular*}{\linewidth}{@{\extracolsep{\fill}}lrrrrrrrr@{}}
    \toprule
    & \multicolumn{4}{c}{sMAPE} & \multicolumn{4}{c}{Dynamics} \\
    \cmidrule(lr){2-5}\cmidrule(l){6-9}
    Model & $H=64$ & $H=128$ & $H=256$ & $H=512$ & $W_1$ & PSD $H^2$ & CD err. & Seg.\ analog \\
    \midrule
    PANDA & 19.09 & 24.42 & 31.94 & 41.22 & 0.588 & 0.195 & 0.537 & 10.03 \\
    \method{}-75M & 16.94 & 22.16 & 28.22 & 35.37 & 0.263 & 0.063 & 0.286 & 5.07 \\
    \method{}-126M & \textbf{16.45} & \textbf{21.52} & \textbf{27.43} & \textbf{34.50} & \textbf{0.251} & \textbf{0.061} & \textbf{0.273} & \textbf{4.80} \\
    \bottomrule
  \end{tabular*}
\end{table}

\begin{figure}[H]
  \centering
  \includegraphics[width=\linewidth]{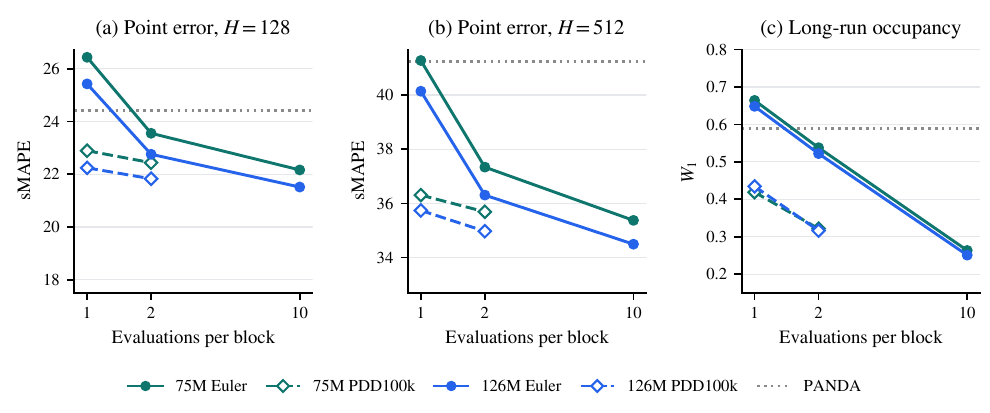}
  \caption{Accuracy versus sampler budget for both model sizes.
  Point errors average all 480 source groups; $W_1$ averages all 135 systems and uses 4,096-step sampled continuations.
  The dotted horizontal line marks PANDA; lower is better.}
  \label{fig:chaos_samplers}
\end{figure}

The sampler comparison evaluates each teacher and its student after 100k PDD updates (\Cref{fig:chaos_samplers,tab:chaos_horizons}).
At one or two evaluations, PDD improves point accuracy over Euler; ten-step Euler remains best within each size at every horizon.
For 126M long continuations, two-evaluation PDD reduces mean $W_1$ from 0.522 to 0.316 relative to two-step Euler, approaching ten-step Euler's 0.251.

\begin{table}[H]
  \centering
  \small
  \setlength{\tabcolsep}{5pt}
  \caption{Mean source-group sMAPE for the Euler teacher and its student after 100k PDD updates.
  NFE counts evaluations per 256-step block and sample; bold marks the lowest error within each size.}
  \label{tab:chaos_horizons}
  \begin{tabular*}{\linewidth}{@{\extracolsep{\fill}}llrrrrr@{}}
    \toprule
    Model & Sampler & NFE & $H=64$ & $H=128$ & $H=256$ & $H=512$ \\
    \midrule
    \multirow{5}{*}{\method{}-75M}
      & Euler & 1 & 20.99 & 26.44 & 32.97 & 41.27 \\
      & Euler & 2 & 17.81 & 23.56 & 30.23 & 37.34 \\
      & Euler & 10 & \textbf{16.94} & \textbf{22.16} & \textbf{28.22} & \textbf{35.37} \\
      & PDD100k & 1 & 17.62 & 22.89 & 29.02 & 36.31 \\
      & PDD100k & 2 & 17.17 & 22.44 & 28.55 & 35.69 \\
    \midrule
    \multirow{5}{*}{\method{}-126M}
      & Euler & 1 & 20.14 & 25.43 & 31.88 & 40.14 \\
      & Euler & 2 & 17.24 & 22.76 & 29.25 & 36.31 \\
      & Euler & 10 & \textbf{16.45} & \textbf{21.52} & \textbf{27.43} & \textbf{34.50} \\
      & PDD100k & 1 & 17.08 & 22.24 & 28.36 & 35.74 \\
      & PDD100k & 2 & 16.69 & 21.83 & 27.86 & 34.97 \\
    \bottomrule
  \end{tabular*}
\end{table}

\FloatBarrier
\Needspace{0.9\textheight}
\subsection{Full continuations}
\label{app:chaos_continuations}

\begin{figure}[!htbp]
  \centering
  \includegraphics[width=\linewidth]{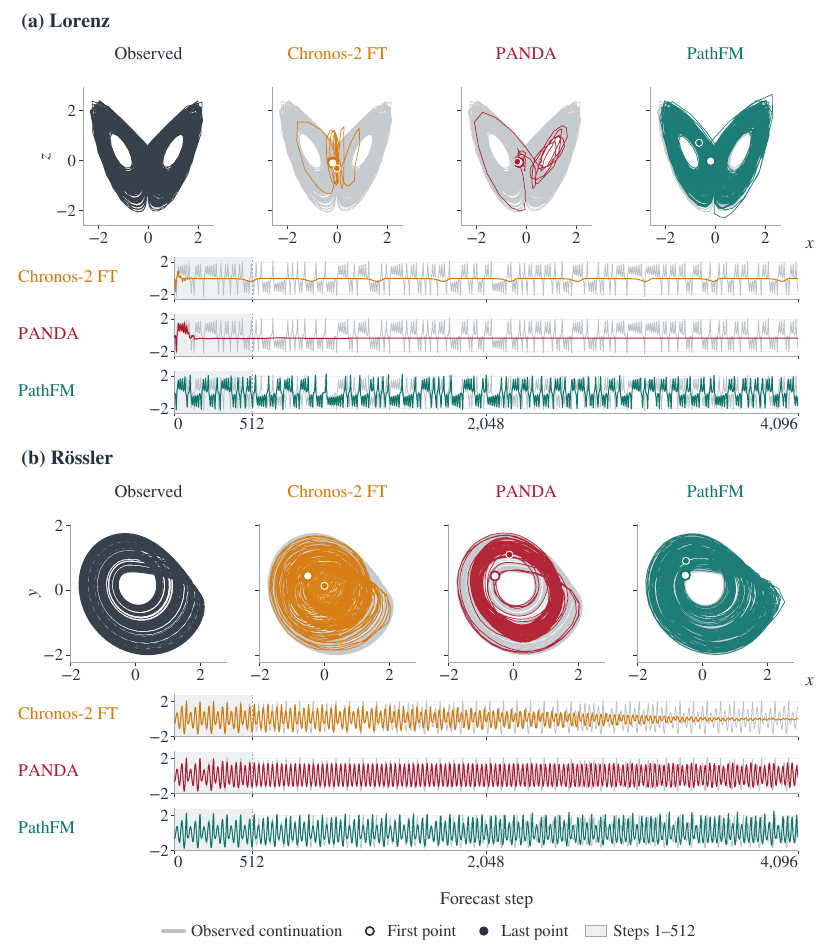}
  \caption{Full 4,096-step Lorenz and R\"ossler continuations.
  All panels use the same initial conditions.
  Gray curves show the observed continuation; circles mark the first and last forecast states, and shading marks the first 512 steps.}
  \label{fig:chaos_full_traces}
\end{figure}

\Cref{fig:chaos_full_traces} extends the qualitative comparison in \Cref{fig:chaos_comparison} to all 4,096 steps and adds R\"ossler.
The displayed \method{}-126M paths use the same checkpoint and ten Euler evaluations per block as the quantitative comparison.
\method{} sustains Lorenz's two-lobe motion and R\"ossler's varying oscillation amplitudes throughout the continuation.
In comparison, \chronos{} and PANDA both collapse to the center of the Lorenz example.
In the R\"ossler example, \chronos{}'s amplitude decays and PANDA settles into more regular oscillations.

\FloatBarrier
\subsection{Metric definitions}
\label{app:chaos_metrics}

\paragraph{Point error.}
With context channel mean $\mu_c$ and population standard deviation $\sigma_c$, let $\widetilde x_{tc}=(x_{tc}-\mu_c)/\max(\sigma_c,10^{-10})$ and define $\widetilde y$ identically.
We use PANDA's factor-100 convention \citep{lai2025panda}:

\begin{equation}
 \mathrm{sMAPE}_h
 =\frac{100}{hC}\sum_{t=1}^{h}\sum_{c=1}^{C}
 \frac{|\widetilde x_{tc}-\widetilde y_{tc}|}
 {|\widetilde x_{tc}|+|\widetilde y_{tc}|+10^{-10}}.
 \label{eq:chaos_smape}
\end{equation}

This scores the blockwise median forecast; individual sampled-path errors are not averaged into the point score.

\paragraph{Wasserstein-1 distance ($W_1$).}
Let $\mathcal R$ be the reference state cloud and $\mathcal S$ the post-burn-in rollout states, standardized using reference channel means and standard deviations.
For $K=64$ seeded unit directions, $\widehat F_{\mathcal R,k}$ and $\widehat F_{\mathcal S,k}$ are the empirical CDFs of the projected states:

\begin{equation}
 \widehat W_1(\mathcal R,\mathcal S)
 =\frac1K\sum_{k=1}^{K}\int_0^1
 \left|\widehat F_{\mathcal R,k}^{-1}(u)
       -\widehat F_{\mathcal S,k}^{-1}(u)\right|\,du.
 \label{eq:sliced_wasserstein}
\end{equation}

The reference cloud retains at most 20,000 states; this calculation subsamples each cloud to at most 10,000.
It measures distance to an empirical state distribution, without an exact invariant measure.

\paragraph{Spectral distance.}
Spectral and segment diagnostics standardize channels using trajectories 0--7.
Welch spectra use 512-point periodic Hann windows, 50\% overlap, unit sampling rate, and constant detrending.
Each spectrum is normalized to unit mass; $p_{x,c}$ is the forecast spectrum and $q_c$ averages the six normalized reference spectra.
The channel-mean squared Hellinger distance is

\begin{equation}
 H^2_{\mathrm{spec}}
 =\frac{1}{2C}\sum_{c=1}^C\sum_f
 \bigl[\sqrt{p_{x,c}(f)}-\sqrt{q_c(f)}\bigr]^2.
 \label{eq:spectral_distance}
\end{equation}

It lies in $[0,1]$ and compares spectral shape independently of total power.

\paragraph{Correlation dimension.}
We estimate $\widehat d_{\mathrm{corr}}$ from log--log slopes of empirical pair-distance probabilities, excluding pairs within a 30-step Theiler window.
The estimator uses at most 4,096 states and 200,000 positive pair distances, taking the median slope over probability ranges $[0.005,0.05]$, $[0.01,0.10]$, and $[0.02,0.20]$, with 20 geometrically spaced probabilities per range.
Against the dataset's stored pooled estimate $\widehat d_{\mathrm{pool}}$, we report

\begin{equation}
 E_{\mathrm{CD}}
 =\left|\widehat d_{\mathrm{corr}}-\widehat d_{\mathrm{pool}}\right|.
 \label{eq:correlation_dimension_error}
\end{equation}

The pooled target uses all ten original trajectories, including the test trajectory, and its own standardization; it is a descriptive geometric reference rather than an independent held-out estimate.

\paragraph{Segment analog.}
Flatten 64-step multivariate segments into vectors in the shared standardized coordinates, without recentering each segment.
Let $\mathcal U_x$, $\mathcal U_{\mathrm{ref}}$, and $\mathcal U_{\mathrm{cal}}$ denote generated, reference, and calibration segments, and $d(u,\mathcal U)=\min_{v\in\mathcal U}\|u-v\|_2$.
The score compares their nearest-reference distances:

\begin{equation}
 A_{\mathrm{seg}}
 =\frac{\operatorname{median}_{u\in\mathcal U_x}d(u,\mathcal U_{\mathrm{ref}})}
 {\max\!\left\{\operatorname{median}_{u\in\mathcal U_{\mathrm{cal}}}
 d(u,\mathcal U_{\mathrm{ref}}),10^{-12}\right\}}.
 \label{eq:segment_analog}
\end{equation}

Generated and calibration segments use stride 32; reference segments use stride eight, retaining at most 20,000.
The calibration normalization gives a reference level of one; larger values indicate less similar local dynamics.

\FloatBarrier

\end{document}